\documentclass[10pt,twocolumn,letterpaper]{article}

\usepackage[pagenumbers]{cvpr} 

\definecolor{cvprblue}{rgb}{0.21,0.49,0.74} 
\usepackage[pagebackref,breaklinks,colorlinks,allcolors=cvprblue]{hyperref}

\newcommand\norm[1]{\lVert#1\rVert_{2}}

\DeclareMathOperator*{\argmin}{arg\,min}
\newcommand{\method}{\textit{PointNSP}\xspace}
\usepackage{algorithm}
\usepackage{algorithmic} 

\usepackage{wrapfig}
\usepackage{multirow}
\usepackage{makecell}

\usepackage{graphicx}

\usepackage{xspace}
\usepackage{amsmath}
\usepackage{amssymb} 
\usepackage{algorithm}
\usepackage{algorithmic}
\usepackage{graphicx}
\usepackage{multirow}
\usepackage{booktabs}
\usepackage{wrapfig}
\usepackage{tcolorbox}
\usepackage{makecell} 
\usepackage{enumitem}
\usepackage{subcaption}
\usepackage{wrapfig}
\usepackage[table]{xcolor}

\def\paperID{15033} 
\def\confName{CVPR}
\def\confYear{2026}

\title{\method: Autoregressive 3D Point Cloud Generation with Next-Scale Level-of-Detail Prediction} 

\author{%
  {
  \textbf{Ziqiao Meng}$^{1}$ \hspace{0.3cm}
  \textbf{Qichao Wang}$^{2}$\hspace{0.3cm}
  \textbf{Zhiyang Dou}$^3$ \hspace{0.3cm}
  \textbf{Zixing Song}$^4$ \hspace{0.3cm}
  \textbf{Zhipeng Zhou}$^2$ \hspace{0.3cm}
  }
   \\
 {
  \textbf{Irwin King}$^5$\hspace{0.35cm}
  \textbf{Peilin Zhao}$^6$ \hspace{0.35cm}
  }
  \vspace{0.35cm}
  \\
  $^1$National University of Singapore $^2$Nanyang Technological University
  $^3$University of Hong Kong \\
  $^4$University of Cambridge 
  $^5$The Chinese University of Hong Kong 
  $^6$Shanghai Jiao Tong University
}

\begin{document}
\maketitle 
\begin{abstract}
Autoregressive point cloud generation has long lagged behind diffusion-based approaches in quality. The performance gap stems from the fact that autoregressive models impose an artificial ordering on inherently unordered point sets, forcing shape generation to proceed as a sequence of local predictions. This sequential bias reinforces short-range continuity but limits the model’s ability to capture long-range dependencies, thereby weakening its capacity to enforce global structural properties such as symmetry, geometric consistency, and large-scale spatial regularities. Inspired by the level-of-detail (LOD) principle in shape modeling, we propose PointNSP, a coarse-to-fine generative framework that preserves global shape structure at low resolutions and progressively refines fine-grained geometry at higher scales through a next-scale prediction paradigm. This multi-scale factorization aligns the autoregressive objective with the permutation-invariant nature of point sets, enabling rich intra-scale interactions while avoiding brittle fixed orderings. Strictly following the baseline experimental setups, empirical results on ShapeNet benchmark demonstrate that PointNSP achieves state-of-the-art (SOTA) generation quality for the first time within the autoregressive paradigm. Moreover, it surpasses strong diffusion-based baselines in parameter, training, and inference efficiency. Finally, under dense generation with $8,192$ points, PointNSP's advantages become even more pronounced, highlighting its strong scalability potential.\footnote{Project Homepage: \url{https://pointnsp.pages.dev}}
\end{abstract}    
\section{Introduction}
\label{sec:intro}

\begin{figure}[!t]
    \centering
    \vspace{-10pt}
    \includegraphics[width=0.43\textwidth]{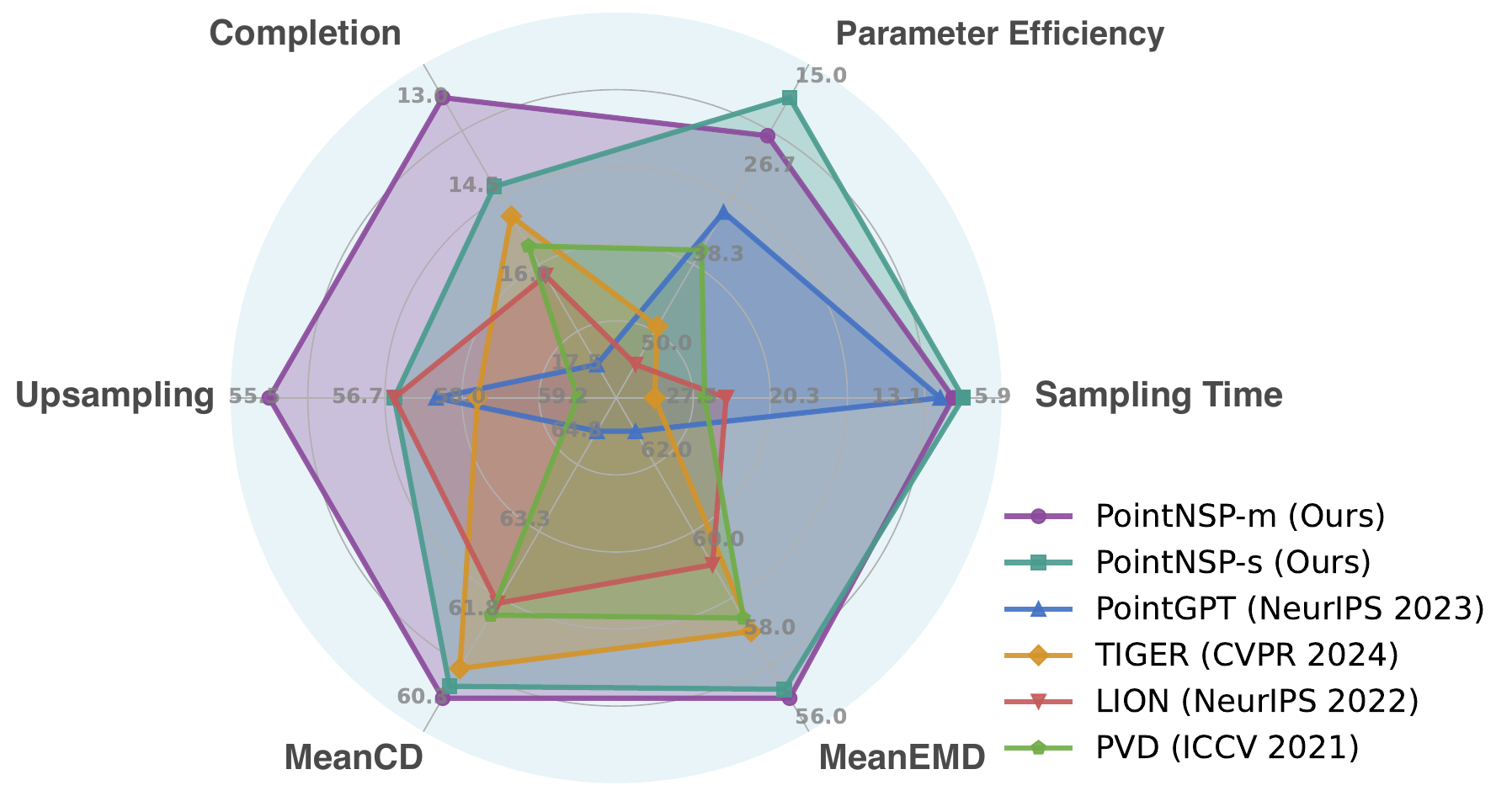} 
    \caption{\method achieves SoTA performance compared to recent strong baseline methods across six key evaluation metrics.}
    \label{fig:intro-comparison}
    \vspace{-10pt}
\end{figure}
Point clouds are a fundamental representation of 3D object shapes, describing each object as a collection of points in Euclidean space. They arise naturally from sensors such as LiDAR and laser scanners, offering a compact yet expressive encoding of fine-grained geometric details. Developing powerful generative models for point clouds is key to uncovering the underlying distribution of 3D shapes, with broad applications in shape synthesis, reconstruction, computer-aided design, and perception for robotics and autonomous systems. However, high-fidelity point cloud generation remains challenging due to the irregular and unordered nature of point sets~\cite{PointNet,PointNet++,DeepSet}. Unlike images or sequences, point clouds lack an inherent ordering—permutations do not alter the underlying shape—rendering naive order-dependent modeling strategies fundamentally misaligned with their structure.

\begin{figure*}[!t]
    \centering
    \includegraphics[width=0.86\textwidth]{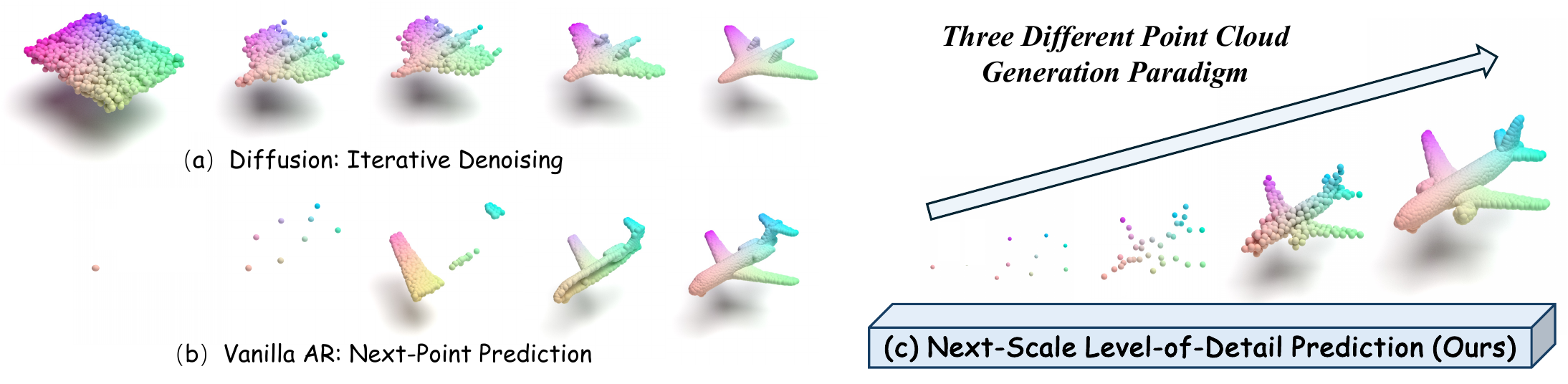}
    \caption{Three types of point cloud generative models: (a) diffusion-based methods that iteratively denoise shapes starting from Gaussian noise; (b) vanilla autoregressive (AR) methods that predict the next point by flattening the 3D shape into a sequence; and (c) our proposed \method, which predicts next-scale level-of-detail in a coarse-to-fine manner.}
    \label{fig:paradigm} 
    \vspace{-10pt}
\end{figure*}

In recent years, diffusion-based methods~\cite{PVD,LION,Tiger} have become the dominant paradigm for 3D point cloud generation, achieving strong empirical performance. However, their Markovian formulation prevents them from leveraging full historical context, often resulting in incoherent shapes. They are also computationally expensive: producing high-quality samples typically requires hundreds to thousands of denoising steps, a burden that becomes prohibitive for dense point clouds. Although advanced samplers~\cite{DDIM,DPM-solver} can accelerate sampling, such speedups commonly degrade generation quality. By contrast, autoregressive (AR) models condition on the entire generation history, which helps enforce local smoothness and generally enables faster sampling. Yet, existing AR approaches must impose an artificial ordering on the inherently unordered point set, typically by flattening it into a sequence, which adheres to a \textit{next-point prediction} paradigm. Point Transformers~\cite{PointTransformer,PointTransformerV2,PointTransformerV3} design specialized architectures for unordered point sets and explore diverse point cloud serialization strategies to improve speed, at the expense of relaxing permutation invariance. PointGrow~\cite{PointGrow} enforces a sequential order by sorting points along the $z$-axis. ShapeFormer~\cite{shapeformer} voxelizes point clouds and flattens codebook embeddings into sequences using a row-major order. PointVQVAE~\cite{pointvqvae} projects patches onto a sphere and arranges them in a spiral sequence to establish a canonical mapping. AutoSDF~\cite{AutoSDF} treats point clouds as randomly permuted sequences of latent variables, while PointGPT~\cite{PointGPT} leverages Morton code ordering to impose structure on unordered data. Although these approaches yield promising results, they still lag behind strong diffusion-based methods~\cite{DPM, PVD, LION, Tiger} in generation quality. This is largely due to the restrictive unidirectional dependencies imposed by fixed sequential orders, which collapse global shape generation into local predictions and inherently violate the fundamental permutation-invariance property. This naturally raises the question: \textit{can we achieve permutation-invariant autoregressive modeling for 3D point cloud generation?}

In this work, we introduce \method, a novel autoregressive framework for 3D point cloud generation that preserves global permutation invariance—a key property ensuring that shapes remain independent of point ordering. \method follows a coarse-to-fine strategy, progressively refining point clouds from global structures to fine-grained details via \textit{next-scale prediction}. Unlike prior approaches that predict one point at a time (next-point prediction), \method captures multiple levels of detail (LoD)~\cite{LoD} at each step, enabling more effective modeling of both global geometry and local structure. This design offers two key advantages. First, it avoids collapsing 3D structures by eliminating the need to flatten point clouds into 1D sequences: each step corresponds to a full 3D shape at a given LoD, ensuring structural coherence and permutation invariance. Second, compared to diffusion-based methods, \method establishes a more structured and efficient generation trajectory, avoiding iterative noise injection and denoising in 3D space. Together, these advances allow \method to achieve high generation quality while maintaining strong modeling efficiency. The comparisons across different paradigms are illustrated in Figure~\ref{fig:paradigm}. 

We conduct extensive experiments on the ShapeNet benchmark to validate the effectiveness of \method across diverse settings. In the standard single-class scenario, \method achieves state-of-the-art (SoTA) \textit{\textbf{generation quality}}, yielding the lowest average Chamfer Distance and Earth Mover’s Distance—setting a new benchmark for autoregressive modeling. Beyond quality, \method also demonstrates substantially higher \textit{\textbf{parameter efficiency}}, \textit{\textbf{training efficiency}}, and \textit{\textbf{sampling speed}} compared to strong diffusion-based baselines. In the more challenging many-class ($55$-class) generation setting, \method maintains SoTA performance, evidencing superior cross-category generalization. Moreover, \method significantly outperforms existing approaches on downstream tasks such as partial point cloud completion and upsampling, further highlighting the robustness and versatility of its design. Comparative results across these metrics are presented in Figure~\ref{fig:intro-comparison}. When evaluated on denser point clouds with $8192$ points, the advantages of \method become even more pronounced, particularly in the aforementioned efficiency metrics, underscoring its \textit{\textbf{scalability potential}}. 

\section{Related Works}
\label{sec:related-works}

\paragraph{Autoregressive Generative Modeling.} The core principle of autoregressive generative models is to synthesize outputs sequentially by iteratively generating intermediate segments. This paradigm has achieved remarkable success in discrete language modeling through next-token prediction~\cite{GPT-3,GPT-4}. Inspired by these advances, researchers have extended autoregressive modeling to other modalities, including images~\cite{VQGAN,RQtransformer,VIT}, speech~\cite{SpeechGPT,SpeechGen}, and multi-modal data~\cite{transfusion,Chameleon}. Although these modalities are often continuous in nature, they are typically transformed into discrete token representations using techniques such as VQ-VAE (VQ)~\cite{VQVAE} or residual vector quantization (RVQ)~\cite{RQtransformer}, with generation performed over the resulting tokens in predefined orders (e.g., raster-scan sequences). To relax the constraint of strict unidirectional dependencies, MaskGIT~\cite{MaskGIT} predicts sets of randomly masked tokens at each step under the control of a scheduler. More recently, VAR~\cite{VAR} redefines the autoregressive paradigm by predicting the next resolution rather than the next token. Within each scale, a bidirectional transformer enables full contextual interaction among tokens, making VAR especially effective for modeling unordered data such as 3D point clouds. Note that while VAR has been applied to 3D mesh generation~\cite{armesh} and 2D image-conditional 3D triplane generation~\cite{SAR3D}, these works differ fundamentally from ours as they operate on different 3D data structures. 

\paragraph{Point Cloud Generation.} Deep generative models have made significant strides in 3D point cloud generation. For instance, PointFlow~\cite{PointFlow} captures the latent distribution of point clouds using continuous normalizing flows~\cite{normalizing-flow, ffjord, neuralode}. Building on this foundation, a series of approaches—including DPM~\cite{DPM}, ShapeGF~\cite{ShapeGF}, PVD~\cite{PVD}, LION~\cite{LION}, TIGER~\cite{Tiger}, PDT~\cite{wang2025pdt}—leverage denoising diffusion probabilistic models~\cite{ddpm, score-sde} to synthesize 3D point clouds through the gradual denoising of input data or latent representations in continuous space. While efforts to improve the sampling efficiency of diffusion models—such as straight flows~\cite{flow-matching, rectified-flow, PSF} and ODE-based solvers~\cite{DDIM}—have achieved acceleration, these methods often introduce trade-offs that compromise generation quality. In contrast, autoregressive models for 3D point cloud generation~\cite{PointGrow, pointvqvae, PointGPT} have received relatively less attention, historically lagging behind diffusion-based techniques in terms of fidelity. In this work, we reformulate point cloud generation as an iterative upsampling process, establishing a connection with sparse point cloud upsampling approaches~\cite{PU-net, Grad-PU, PUDM}. A comprehensive review of point cloud upsampling works is provided in the Appendix~\ref{appendix:more-related-works}. 
\section{Method: PointNSP}
\label{sec:method}

\subsection{Preliminaries: Autoregressive 3D Point Cloud Generation} 
A point cloud is represented as a set of $N$ points $\mathbf{X} = \lbrace \mathbf{x}_{i}\rbrace_{i=1}^{N}$, where each point $\mathbf{x}_{i}\in \mathbb{R}^{3}$ corresponds to a 3D coordinate. Prior autoregressive approaches~\citep{pointvqvae,PointGrow,PointGPT} mainly follow the next-point prediction paradigm:
\begin{equation}
\label{eq:ntp}
    p(\mathbf{x}_{1},\mathbf{x}_{2},\dots,\mathbf{x}_{N}) = \prod_{i=1}^{N}p(\mathbf{x}_{i}|\mathbf{x}_{i-1},\dots,\mathbf{x}_{2},\mathbf{x}_{1}). 
\end{equation}
Training Eq.~\ref{eq:ntp} requires a sequential classification objective. Each point $\mathbf{x}_{i}$ is first converted into a discrete integer token $q_{i}$ via VQ-VAE quantization~\citep{VQVAE}, and the resulting tokens are flattened into a sequence $(q_{1}, \dots, q_{N})$ according to a \textit{predefined generation order}. Autoregressive modeling is then applied to this discrete sequence following the paradigm in Eq.~\ref{eq:ntp}. However, this approach struggles to preserve permutation invariance, since the probability distribution depends on the chosen token ordering and is not invariant to different permutations of the points. 
\begin{figure*}[ht!]
    \centering
    \includegraphics[width=0.95\textwidth]{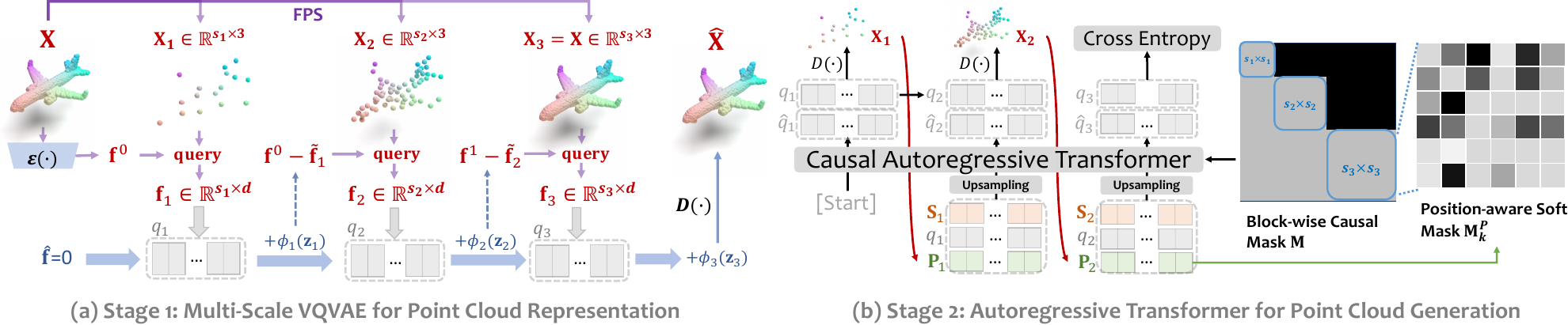}
    \caption{(a) Illustration of training a multi-scale VQVAE in a residual manner for point cloud representation across scales $s_{1}$ to $s_{3}$, resulting in a multi-scale token sequence $Q = (q_{1}, \dots, q_{3})$; (b) Illustration of training a causal transformer with intermediate shape decoding, scale token $\operatorname{upsampling}$ ($s_{1}\rightarrow s_{2}$ and $s_{2}\rightarrow s_{3}$), position-aware soft masks $\mathbf{M}^{P}_{k}$, and block-wise causal masks $\mathbf{M}$.}
    \label{fig:stage-1} 
    \vspace{-10pt}
\end{figure*}
In this work, instead of predicting the next point, we propose a novel autoregressive framework that predicts the \textit{next-scale level-of-detail (LoD)} of the point cloud $\mathbf{X}$, while simultaneously preserving the permutation invariance property: 
\begin{equation}
\label{eq:permutation-invariance}
    p(\pi(\mathbf{x}_1,\dots,\mathbf{x}_N)) = p(\mathbf{x}_1,\dots,\mathbf{x}_N), \qquad\forall \ \pi \in S_N.
\end{equation}
In general, we construct $k$ different LoDs of $\mathbf{X}$, forming a coarse-to-fine sequence of global shapes ${\mathbf{X}_{1}, \dots, \mathbf{X}_{K}}$, where each $\mathbf{X}_{k} \in \mathbb{R}^{s_{k} \times 3}$ represents a global shape at resolution $s_{k}$, obtained by downsampling the original $\mathbf{X}$. Then, \method is designed to learn the following distribution: 
\begin{equation}
\label{eq:nsp}
    p(\mathbf{X}_{1},\mathbf{X}_{2},\dots,\mathbf{X}_{K}) = \prod_{k=1}^{K}p(\mathbf{X}_{k}|\mathbf{X}_{k-1},\dots,\mathbf{X}_{2},\mathbf{X}_{1}). 
\end{equation}
The generation process bears a strong resemblance to an autoregressive upsampling procedure, governed by a sequence of upsampling rates $r_{1},r_{2}, \dots, r_{K-1}$ that satisfy the relation $r_{K-1}\times \dots \times r_{1}\times s_{1} = s_{K}$. We set $s_{K}=N, \mathbf{X}_{K}=\mathbf{X}$ to reconstruct the shape at the target full-resolution point count. Under this formulation, Eq.~\ref{eq:nsp} requires learning a hierarchy of representations $\mathbf{X}_{k}$ across multiple LoD information. These representations progressively encode more global and semantically coherent structural information than the next-point prediction in Eq.~\ref{eq:ntp}. 

\subsection{Multi-Scale LoD Representation} 
\label{subsec:multi-scale-LoD}

\paragraph{Sampling LoD Sequence.} 
To construct the LoD sequence $(\mathbf{X}_{1}, \dots, \mathbf{X}_{K})$ for each sample $\mathbf{X}$, three key considerations must be addressed. First, the sampling operation should preserve the permutation invariance property of point clouds, ensuring that the resulting LoD sequence remains independent of the input point ordering in $\mathbf{X}$. Second, the sampling strategy should aim for \textit{comprehensive spatial coverage} at each LoD, reflecting the existence of an underlying continuous surface that the 3D shape represents. To this end, we adopt the Farthest Point Sampling (FPS)~\citep{FPS} algorithm to construct the LoD sequence. The point selection mechanism in FPS relies exclusively on spatial geometry—specifically, pairwise Euclidean distances—rather than the ordering of points within the input set, thereby preserving permutation invariance. Moreover, the inherent stochasticity introduced by the random initialization of the starting point enables diverse subsets $\mathbf{X}_{k}$ to be sampled at each scale $k$ across different training epochs, thereby enhancing the diversity and representational robustness of the generated LoD sequences. 

\paragraph{Multi-Scale Feature Extraction.} Rather than directly learning tokenizers from the sampled point clouds in the 3D coordinate space, we learn their corresponding quantized representations within the latent feature space defined by the LoD sequence. To obtain latent features $\mathbf{f}^{0} \in \mathbb{R}^{N\times d}$ from the point cloud $\mathbf{X}$, any permutation-equivariant network $\text{NN}(\cdot)$ is applicable: the per-point features reorder consistently with any permutation of the input points, i.e., $\pi(\text{NN}(\mathbf{x}_{1},...,\mathbf{x}_{N})) = \text{NN}(\pi(\mathbf{x}_{1},...,\mathbf{x}_{N}))$ for any permutation $\pi$. Therefore, permutation-equivariant architectures such as PointNet~\citep{PointNet}, PointNet++~\citep{PointNet++}, PointNeXt~\citep{PointNext} and PVCNN~\citep{PV-CNN} are all applicable. To encourage each scale to capture \textit{complementary information} rather than redundantly encoding features already represented at coarser levels, we extract latent features $\mathbf{f}_{k}\in \mathbb{R}^{s_{k}\times d}$ in a residual fashion: 
\begin{equation}
\label{eq:residual-query}
    \mathbf{f}_{k} = \operatorname{query}(\mathbf{f}^{k-2} - \tilde{\mathbf{f}}_{k-1}, \mathbf{X}_{k}),\ \mathbf{f}_{1} = \operatorname{query}(\mathbf{f}^{0}, \mathbf{X}_{1}).
\end{equation}
Here, $\tilde{\mathbf{f}}_{k-1}$ represents the feature contribution from the learned tokenizers at scale $k-1$, while $\operatorname{query}(\cdot)$ retrieves latent features according to the index correspondence between the sampled subset $\mathbf{X}_{k}$ and the original point set $\mathbf{X}$, ensuring consistent alignment across scales. This produces a sequence of LoD latent features $(\mathbf{f}_{1},\dots, \mathbf{f}_{K})$. 

\paragraph{Multi-Scale VQVAE Tokenizer.} For each scale $k$, we learn tokenizers for latent feature $\mathbf{f}_{k}$ through quantization $\mathcal{Q}$ into discrete tokens $q_{k} = (q^{1}_{k}, q^{2}_{k},\dots, q^{s_k}_{k})  = \mathcal{Q}(\mathbf{f}_{k})\in [V]$, where $[V]$ denotes a sequence of entries corresponding to indices in a learnable codebook $Z \in \mathbb{R}^{V\times d}$ containing $V$ vectors. Note that this codebook $Z$ is shared across all scales for efficient utilization. Each token $q^{i}_{k}$ indexes the nearest codebook embedding $\mathbf{z}_{v}\in \mathbb{R}^{d}$ to the corresponding latent feature $\mathbf{f}_{k}[i] \in \mathbb{R}^{d}$: $q^{i}_{k}= \argmin_{v\in [V]}\norm{\mathbf{z}_{v}-\mathbf{f}_{k}[i]}$. This produces scale-wise token sequence $Q = (q_{1},\dots,q_{K})$ along with their corresponding embeddings $(\mathbf{z}_{1}\in \mathbb{R}^{s_{1}\times d},\dots,\mathbf{z}_{K}\in \mathbb{R}^{s_{K}\times d})$, forming a hierarchical discrete representation aligned with the LoD sequence. For each scale $k$, the feature contribution $\tilde{\mathbf{f}}_{k}$ produced by the scale-specific tokenizers $q_{k}$ is then given by:
\begin{equation} 
\label{eq:feature-contribution}
\tilde{\mathbf{f}}_{k} = \phi_{k}(\operatorname{upsampling}(\mathbf{z}_{k}, s_{K})),\ \mathbf{z}_{k} = \operatorname{lookup}(Z, q_{k}),
\end{equation}
where $\phi_{k}(\cdot): \mathbb{R}^{N\times d}\rightarrow \mathbb{R}^{N\times d}$ is a permutation-equivariant network that refines the latent embeddings, and $\operatorname{upsampling}(\cdot, s_{K})$ increases the resolution of the latent $\mathbf{z}_{k}\in \mathbb{R}^{s_{k}\times d}$ to the highest resolution $s_{K} \times d$. 

\paragraph{Upsampling \& Reconstruction.} The partial sum of all feature contributions is computed as $\hat{\mathbf{f}} = \sum_{k=1}^{K}\tilde{\mathbf{f}}_{k}$, and the final predicted 3D shape $\mathbf{\hat{X}}$ is reconstructed via a simple MLP decoder $D(\cdot): \mathbb{R}^{N\times d} \rightarrow \mathbb{R}^{N\times 3}$: $\hat{\mathbf{X}}=D(\hat{\mathbf{f}})$. This formulation demonstrates that $\hat{\mathbf{X}}$ is obtained by aggregating information across all scales of the LoD hierarchy, effectively combining coarse and fine features to generate the final 3D reconstruction. The $\operatorname{upsampling}$ in Eq.~\ref{eq:feature-contribution} follows a PU-Net~\cite{PU-net}-inspired procedure, which consists of $\operatorname{duplication}$ and $\operatorname{reshaping}$ operations:
\begin{equation}
\label{eq:up-sample}
    \mathbf{z}_{k} (s_{k}\times d) \xrightarrow{\operatorname{duplicate}} \mathbf{z}_{k} (s_{k}\times r \times d) \xrightarrow{\operatorname{reshape}} \mathbf{z}_{k} ((s_{k}\cdot r) \times d),
\end{equation}
where $\mathbf{z}_{k}\in \mathbb{R}^{(s_{k}\cdot r) \times d} = \mathbf{z}_{K}\in \mathbb{R}^{s_{K} \times d}$ denotes the upsampled representation at the largest scale $s_{K}$, with upsampling rate $r = \frac{s_{K}}{s_{k}}$ specified in advance. This operation densifies points by arbitrary factors while preserving the permutation-equivariance of the latent features. The reconstruction loss is defined as: 
\begin{equation}
    \mathcal{L}_{\text{recon}} = \mathcal{L}_{\text{CD}}(\mathbf{X}, \hat{\mathbf{X}}) + \mathcal{L}_{\text{EMD}}(\mathbf{X}, \hat{\mathbf{X}}) + \sum_{k=1}^{K}||\mathbf{f}_{k} - sg(\mathbf{z}_{k})||_2^2,
\end{equation}
where $\mathcal{L}_{\text{CD}}$ and $\mathcal{L}_{\text{EMD}}$ denote the Chamfer Distance and Earth Mover’s Distance losses, commonly used to evaluate point cloud similarity from complementary perspectives. The stop-gradient operation $sg[\cdot]$ ensures that the latent features $\mathbf{f}_{k}$ used for reconstruction remain consistent with the quantized latent vectors $\mathbf{z}_{k}$. The above process is described in Algorithm~\ref{alg:encoder} and ~\ref{alg:decoder} and Figure~\ref{fig:stage-1} (a). 

\begin{algorithm}
   \caption{Multi-scale VQVAE encoder} 
   \label{alg:encoder} 
\begin{algorithmic}[1]
   \STATE {\bfseries Input:} 3D point cloud $\mathbf{X} = \lbrace \mathbf{x}_{i}\rbrace_{i=1}^{N}$. 
   \STATE {\bfseries Hyperparameters:} \# scales $\lbrace s_{k}\rbrace_{k=1}^{K}$. 
   \STATE $\mathbf{f}^{0} = \mathcal{E}(\mathbf{X})$, $Q = []$;
   \FOR{$k=1,\cdots,K$} 
    \STATE $\mathbf{X}_k = \operatorname{FPS}(\mathbf{X}, s_{k})$\
    \STATE $q_k = \mathcal{Q}(\mathbf{f}_{k}), \mathbf{f}_k = \operatorname{query}(\mathbf{f}^{k}, \mathbf{X}_{k})$\;
    \STATE $Q = \operatorname{queue\_push}(Q, q_k)$\; 
    \STATE $\mathbf{z}_k = \operatorname{lookup}(Z, q_k)$\; 
    \STATE $\mathbf{z}_k = \operatorname{upsampling}(\mathbf{z}_k, s_K)$\; 
    \STATE $\mathbf{f}^{k} = \mathbf{f}^{k-1} - \phi_k(\mathbf{z}_k)$\; 
   \ENDFOR 
   \STATE {\bfseries Return:} token sequence $Q = (q_{1},\dots,q_{K})$.\
\end{algorithmic}
\end{algorithm}
\begin{algorithm}
   \caption{Multi-scale VQVAE decoder} 
   \label{alg:decoder}
    \begin{algorithmic}[1]
   \STATE {\bfseries Input:} token sequence $Q = (q_{1},\dots,q_{K})$.\; 
   \STATE {\bfseries Hyperparameters:} \# scales $\lbrace s_{k}\rbrace_{k=1}^{K}$.
   \STATE $\hat{\mathbf{f}} = 0 $\;
   \FOR{$k=1,\cdots,K$} 
    \STATE $q_k = \operatorname{queue\_pop}(Q)$\;
    \STATE $\mathbf{z}_k = \operatorname{lookup}(Z, q_k)$\; 
    \STATE $\mathbf{z}_k = \operatorname{upsampling}(\mathbf{z}_k, s_K)$\;
    \STATE $\hat{\mathbf{f}} = \hat{\mathbf{f}} + \phi_k(\mathbf{z}_k)$\;
   \ENDFOR
   \STATE $\hat{\mathbf{X}} = \mathcal{D}(\hat{\mathbf{f}}) $\;
   \STATE {\bfseries Return:} reconstructed point cloud $\hat{\mathbf{X}}$\;
\end{algorithmic}
\end{algorithm}

\subsection{Autoregressive Transformer for Next-Scale LoD Prediction}

The next step is to train an autoregressive transformer on the input multi-scale token sequence $Q = (\operatorname{[start]}, q_{1},\dots,q_{K-1})$ to predict $(q_{1},\dots,q_{K})$. Owing to the strong local-geometry inductive bias in 3D structures, a standard causal transformer struggles to capture both intra-scale and inter-scale dependencies efficiently. Therefore, it is necessary to incorporate geometric information into the transformer design, which poses additional challenges due to the unordered nature of point clouds. 

\paragraph{Inter-Scale Interaction Modeling.} Inter-scale interactions across input scales $(q_{1},\dots,q_{K-1})$ are critical for the model to capture dependencies between levels of detail and to generate each subsequent LoD conditioned on the preceding ones. We follow a key principle: \textit{Tokens at scale $q_{k}$ are permitted to attend only to tokens from preceding scales $q_{1},\dots, q_{k-1}$. Within each scale, however, all tokens $q_{k} = (q^{1}_{k},\dots, q^{s_{k}}_{k})$ are allowed to fully attend to one another, ensuring that the model interprets them as complete shapes at the corresponding resolution.} To enforce this constraint, we construct a causal mask $\mathbf{M}\in \mathbb{R}^{(s_{1}+\dots+s_{K})\times(s_{1}+\dots+s_{K})}$ as a block-diagonal matrix, where each diagonal block $\mathbf{M}_{k}$ of size $s_{k}\times s_{k}$ is fully unmasked: $\mathbf{M} = \operatorname{diag}[\mathbf{M}_{1}, \mathbf{M}_{2}, ..., \mathbf{M}_{K}]$. This design allows bidirectional attention within each scale while maintaining a lower-triangular dependency structure across scales, thereby preventing information leakage from future scale tokens. To further distinguish scales, each token is assigned a scale embedding $\mathbf{s}^{i}_{k}\in \mathbb{R}^{d}$, implemented as a one-hot embedding over $K$ scales and shared among tokens within the same scale. Algorithmic details are explained in the Appendix~\ref{appendix:details}. 

\paragraph{Intra-Scale Interaction Modeling.} Since the intra-scale bidirectional transformer does not inherently encode positional information, stacking multiple transformer layers can dilute the relative positional signals among tokens. To mitigate this issue, we first add positional encodings to the token embeddings after each transformer layer. In addition, we augment $\mathbf{M}_{k}$ with a position-aware soft masking matrix $\mathbf{M}^{p}_{k} \in \mathbb{R}^{s_{k}\times s_{k}}$, which is derived from the coordinate-based absolute positional embedding matrix $\mathbf{P}_{k}\in \mathbb{R}^{s_{k}\times d}$: 
\begin{equation}
\label{eq:p-attention} 
    \mathbf{M}^{p}_{k} = \operatorname{Softmax}((\mathbf{P}_{k}\mathbf{W}_{p})(\mathbf{P}_{k}\mathbf{W}_{p})^{T}),\ \mathbf{W}_{p}\in \mathbb{R}^{d\times d}.
\end{equation} 
$\mathbf{M}^{p}_{k}$ is a symmetric matrix, where each entry $\mathbf{M}^{p}_{k}[i,j]\in (0,1)$ encodes the soft relative position information between points $i$ and $j$. This raises a key question: \textit{how can we derive the positional embedding $\mathbf{P}_{k}$ for each scale when no explicit 3D geometry is available at this stage}? Our solution is an intermediate-structure decoding strategy. Specifically, we apply the decoder $D(\cdot)$ to reconstruct the intermediate structure $\mathbf{X}_{k}$ using all ground-truth tokens up to step $k$:
\begin{equation}
\label{eq:intermediate}
    \mathbf{X}_{k} = D(\sum_{m=1}^{k}\phi_{m}(\operatorname{upsampling}(\mathbf{z}_{m}, s_{m}))), 
\end{equation}
where $(\mathbf{z}_{1}, \dots, \mathbf{z}_{k}) = \operatorname{lookup}(Z, (q_{1}, \dots, q_{k}))$, $\mathbf{X}_{k}\in \mathbb{R}^{s_{k}\times 3}$ provides the coordinate information used to compute the positional embedding $\mathbf{P}_{k}$. $\mathbf{P}_{k}$ is derived as an absolute positional encoding based on the 3D coordinates $\mathbf{X}_{k}$ using trigonometric functions (e.g. $\sin$ and $\cos$). During inference, the predicted token $\hat{q}_{k}$ is used to obtain $\mathbf{\hat{z}}_{k}$ in Eq.~\ref{eq:intermediate}, rather than the ground-truth token $q_{k}$, which is not available at test time. The detailed derivation of $\mathbf{P}$ is provided in the Appendix~\ref{appendix:details}. Note that any positional encoding based on token indices should not be applied, as it would violate the permutation equivariance property.

The prediction of each token $\hat{q}_{k}^{i}$ is evaluated using the cross-entropy (CE) loss $\mathcal{L}^{i}_{k}=\operatorname{CE}(\hat{q}_{k}^{i}, q^{i}_{k})$. We first compute intra-scale loss $\mathcal{L}_{k}=\frac{1}{s_{k}}\sum_{i=1}^{s_{k}}\mathcal{L}^{i}_{k}$ and then compute inter-scale loss $\mathcal{L}_{\text{total}} = \frac{1}{K}\sum_{k=1}^{K}\mathcal{L}_{k}$. The second stage training architecture is illustrated in Figure~\ref{fig:stage-1} (b). We provide a theoretical analysis of the permutation invariance of \method’s distribution modeling in Appendix~\ref{appendix:PI}.

\label{sec:exp}
\begin{table*}[!htbp]
    \centering
     \resizebox{0.99\textwidth}{!}{
     \begin{tabular}{l|c|cc|cc|cc|c|c}
\Xhline{3\arrayrulewidth} 
      \multirow{2}{*}{\textbf{Model}} & \multirow{2}{*}{\textbf{Generative Model}} & \multicolumn{2}{c|}{\textbf{Airplane}} & \multicolumn{2}{c|}{\textbf{Chair}} & \multicolumn{2}{c|}{\textbf{Car}} & \multirow{2}{*}{\textbf{Mean CD $\downarrow$}} & \multirow{2}{*}{\textbf{Mean EMD $\downarrow$}}\\
      \cline{3-8}
      & & CD $\downarrow$ & EMD $\downarrow$ & CD $\downarrow$ & EMD $\downarrow$ & CD $\downarrow$ & EMD $\downarrow$ & & \\
    \Xhline{1.5\arrayrulewidth} 
       ShapeGF~\citep{ShapeGF} & Diffusion & 80.00 & 76.17 & 68.96  & 65.48 & 63.20 & 56.53 & 70.72 & 66.06\\
       DPM~\citep{DPM} & Diffusion & 76.42 & 86.91 & 60.05 & 74.77 & 68.89 & 79.97 & 68.45 & 80.55\\
       PVD~\citep{PVD} & Diffusion & 73.82 & 64.81 & 56.26 & 53.32 & 54.55 & 53.83 & 61.54 & 57.32 \\
       LION~\citep{LION} & Diffusion & 72.99 & 64.21 & 55.67 & 53.82 & 53.47 & 53.21 & 61.75 & 57.59\\
       TIGER~\citep{Tiger} & Diffusion & 73.02 & 64.10 & 55.15 & 53.18 & 53.21 & 53.95 & 60.46 & 57.08\\ 
        PointGrow~\citep{PointGrow} & Autoregressive  & 82.20 & 78.54 & 63.14 & 61.87 & 67.56 & 65.89 & 70.96 & 68.77\\
        CanonicalVAE~\citep{pointvqvae} & Autoregressive & 80.15 & 76.27 & 62.78 & 61.05 & 63.23 & 61.56 & 68.72 & 66.29\\
        PointGPT~\citep{PointGPT} & Autoregressive & 74.85 & 65.61 & 57.24 & 55.01 & 55.91 & 54.24 & 63.44 & 62.24\\ 
        \method-s (\textit{ours}) & Autoregressive & \underline{72.92} & \underline{63.98} & \underline{54.89} & \underline{53.02} & \underline{52.86} & \underline{52.07} & \underline{60.22} & \underline{56.36}\\
        \rowcolor{green!20}
        \textbf{\method-m (\textit{ours})} & \textbf{Autoregressive} & \textbf{72.24} & \textbf{63.69} & \textbf{54.54} & \textbf{52.85} & \textbf{52.17} & \textbf{51.85} & \textbf{59.65} & \textbf{56.13}\\
 
 \midrule 
    LION & Diffusion & 67.41 & 61.23 & \underline{53.70} & 52.34 & \underline{53.41} & 51.14 & \underline{58.17} & 54.90\\
    TIGER & Diffusion & 67.21 & 56.26 & 54.32 & 51.71 & 54.12 & 50.24 & 58.55 & 52.74\\
    \method-s (\textit{ours}) & Autoregressive & \underline{67.15} & \underline{56.12} & 54.22 & \underline{51.19} & 53.98 & \underline{50.15} & 58.45 & \underline{52.49} \\
    \rowcolor{green!20}
    \textbf{\method-m (\textit{ours})} & \textbf{Autoregressive} & \textbf{66.98} & \textbf{56.05} & \textbf{54.01} & \textbf{53.76} & \textbf{53.12} & \textbf{50.09} & \textbf{58.04} & \textbf{52.30}\\
    \Xhline{3\arrayrulewidth} 
    \end{tabular}
    }
    \caption{Performance under the standard $2048$-point setup on ShapeNet is reported for two dataset splits: the top corresponds to the conventional random split, and the bottom corresponds to the LION split~\cite{LION}. The best results are highlighted in bold with a green bar, and the second-best results are underlined.} 
    \label{tab:main_result}
\end{table*}

\begin{figure*}[ht!]
    \centering
    \includegraphics[width=0.95\textwidth]{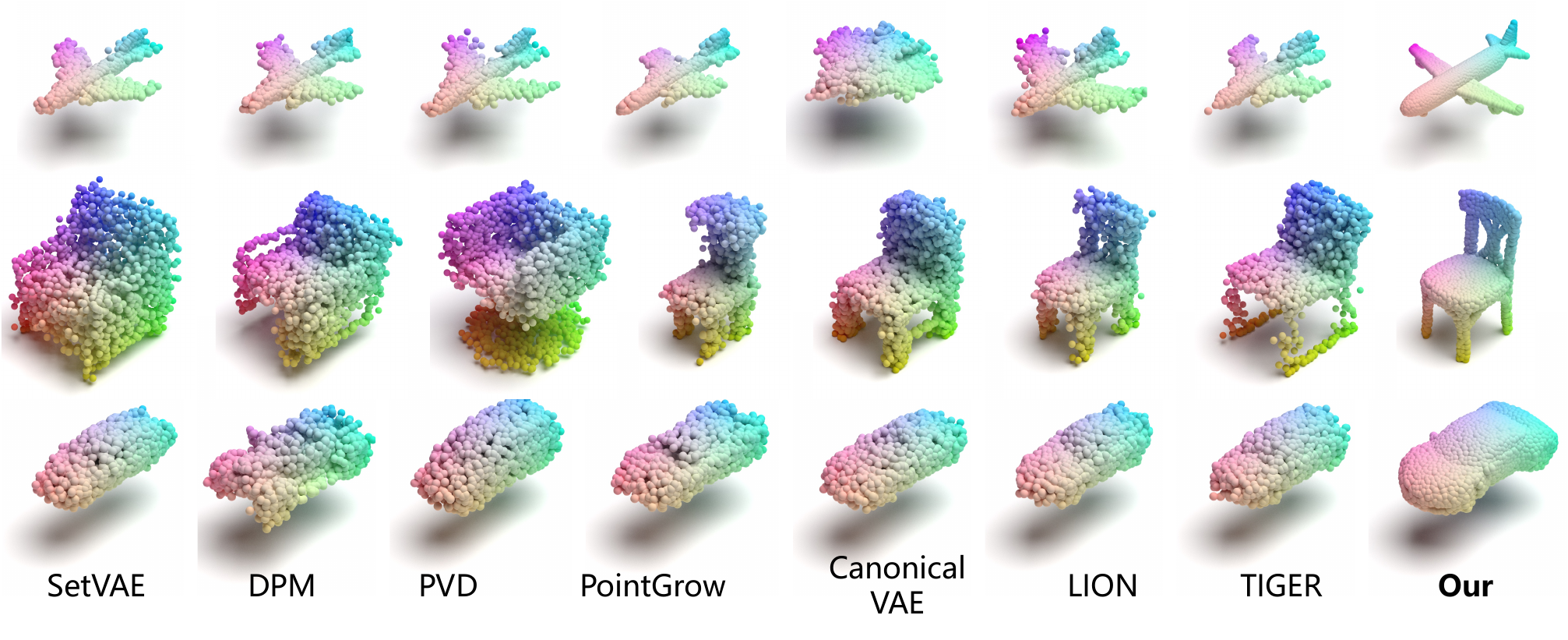}
    \caption{Visualization of generation results compared with baseline models. \method produces high-quality and diverse 3D point clouds.} 
    \label{fig:visualization}
\end{figure*}
\section{Experiments}
\label{sec:experiments}

\begin{figure*}[ht!]
    \centering
    \includegraphics[width=0.9\textwidth]{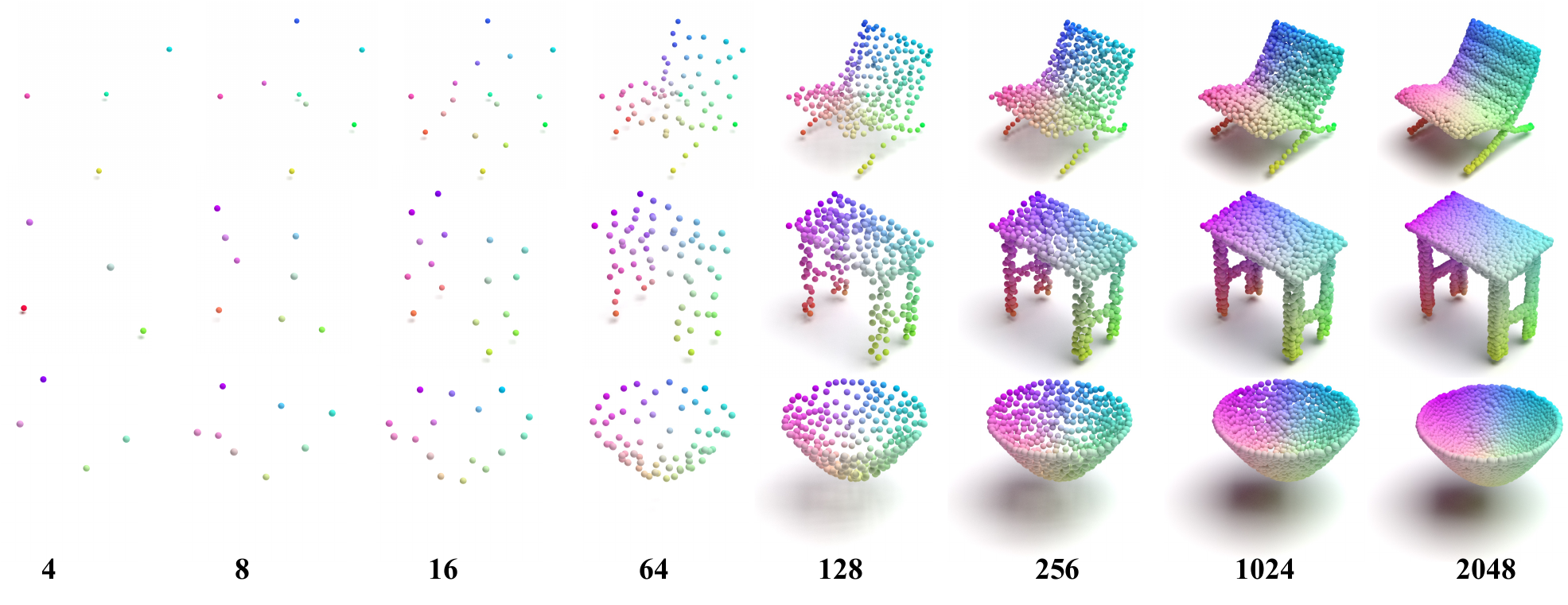}
    \caption{Visualization of multi-scale point clouds during the \method generation process as the scale $K$ increases.}
    \label{fig:visualization_multi_scale}
\end{figure*}

\begin{table*}[ht]
\centering
\resizebox{0.99\textwidth}{!}{
\begin{tabular}{l|cc|cc|cc|cc|cc|cc}
\Xhline{3\arrayrulewidth}
\multirow{3}{*}{\textbf{Model}} & \multicolumn{6}{c|}{\textbf{Dense Point Cloud Generation (8192 pts)}} & \multicolumn{6}{c}{\textbf{Many-Class Generation (55-Class)}} \\
\cline{2-13}
& \multicolumn{2}{c|}{\textbf{Airplane}} & \multicolumn{2}{c|}{\textbf{Chair}} & \multicolumn{2}{c|}{\textbf{Car}} &
  \multicolumn{2}{c|}{\textbf{Airplane}} & \multicolumn{2}{c|}{\textbf{Chair}} & \multicolumn{2}{c}{\textbf{Car}} \\
& CD $\downarrow$ & EMD $\downarrow$ & CD $\downarrow$ & EMD $\downarrow$ & CD $\downarrow$ & EMD $\downarrow$ &
  CD $\downarrow$ & EMD $\downarrow$ & CD $\downarrow$ & EMD $\downarrow$ & CD $\downarrow$ & EMD $\downarrow$ \\
\Xhline{3\arrayrulewidth}
PVD & 69.77 & 69.98 & 52.56 & \underline{51.33} & 54.19 & 46.55 & 97.53 & 99.88 & 88.37 & 96.37 & 89.77 & 94.89 \\
TIGER & 68.48 & 60.24 & 51.87 & 51.85 & 53.78 & 44.12 & 83.54 & 81.55 & \underline{57.34} & 61.45 & 65.79 & 57.24 \\
PointGPT & 69.29 & 61.56 & 52.43 & 52.09 & 54.97 & 45.20 & 94.94 & 91.73 & 71.83 & 79.00 & 89.35 & 87.22 \\
LION & 68.95 & 60.38 & 53.76 & 53.45 & 54.98 & 44.67 & 86.30 & 77.04 & 66.50 & 63.85 & 64.52 & 54.21 \\
\cline{1-13}
\textit{\method-s} (\textit{ours}) & \underline{67.84} & \underline{58.25} & \underline{51.26} & 51.81 & \underline{52.40}  & \underline{43.76} & \underline{78.95} & \underline{68.84} & 58.79 & \underline{55.10} & \underline{59.97} & \underline{52.89} \\
\rowcolor{green!20}
\textbf{\method-m (\textit{ours})} & \textbf{66.63} & \textbf{55.29} & \textbf{50.98} & \textbf{50.45} & \textbf{52.12} & \textbf{43.05} & \textbf{75.42} & \textbf{66.54} & \textbf{56.03} & \textbf{52.22} & \textbf{57.95} & \textbf{49.55} \\
\Xhline{3\arrayrulewidth}
\end{tabular}}
\caption{Comparison of dense point cloud generation (left) and many-class generation (right). CD and EMD metrics ($\downarrow$) are reported.}
\label{tab:merged_dense_manyclass}
\end{table*}

\subsection{Experimental Setup}

\paragraph{Datasets \& Metrics.} In line with prior studies, we adopt ShapeNetv2, pre-processed by PointFlow~\cite{PointFlow}, as our primary dataset. Each shape is globally normalized and uniformly sampled to $2048$ points (standard setting). Experiments under this setting are conducted on two data splits: the standard random split and the LION split~\cite{LION,Tiger}. We further evaluate a dense generation setting with $8192$ points (dense setting). Following established benchmarks~\cite{PVD,LION}, we use the 1-nearest neighbor (1-NN) accuracy~\cite{1-NN} as our primary evaluation metric, which effectively captures both the quality and diversity of generated point clouds—a score near $50\%$ indicates strong performance~\cite{PointFlow}. The 1-NN distance matrix is computed using two widely adopted distance measures: Chamfer Distance (CD) and Earth Mover’s Distance (EMD). We also report mean CD and mean EMD, averaged across all object categories. To assess efficiency, we record the total training time in GPU hours and report both inference time and parameter count.

\begin{table}[ht]
    \centering
    \scalebox{0.55}{
     \begin{tabular}{c|l|cc|cc|c}
    \Xhline{4\arrayrulewidth} 
      \multirow{2}{*}{\textbf{Quality}$\downarrow$} &\multirow{2}{*}{\textbf{Model}} &
      \multicolumn{2}{c|}{\textbf{2048 pts}} &
      \multicolumn{2}{c|}{\textbf{8192 pts}} &
      \multirow{2}{*}{\textbf{Param $\downarrow$}} \\
      \cline{3-6}
        & & \textbf{Training}$\downarrow$ & \textbf{Sampling}$\downarrow$ &
        \textbf{Training}$\downarrow$ & \textbf{Sampling}$\downarrow$ & \\
      \cline{1-7}
      6 & PointGPT  & 185 & 5.32 & 296 & 10.56 & 46\\
      5 & PVD  & \underline{142} & 29.9 & 201 & 58.1 & 45\\
      4 & LION  & 550 & 31.2 & 610 & 59.5 & 60\\
      3 & TIGER  & 164 & 23.6 & 320 & 42.1 & 55\\
      \cline{1-7}
      \rowcolor{green!20}
      2 & \textbf{\method-s (\textit{ours})} & \textbf{125} & \textbf{3.21} & \textbf{175} & \textbf{4.54} & \textbf{22} \\ 
      1 & \method-m (\textit{ours}) & 178 & \underline{3.59} & \underline{190} & \underline{5.48} & \underline{32} \\ 
    \Xhline{3\arrayrulewidth}
    \end{tabular}}
    \caption{Training time (in GPU hours, averaged over three categories), sampling time (in seconds, averaged over samples), and model size (in millions of parameters). Ranked by generation quality on $2048$ and $8192$ settings.} 
    \label{tab:traing-inference-time}
\end{table}

\subsection{Single-Class Generation: Standard \& Dense} 

\paragraph{Results.}  
To comprehensively assess the performance of \method, we compare it with several strong baseline models. Specifically, we include SoTA diffusion-based approaches—ShapeGF~\cite{ShapeGF}, DPM~\cite{DPM}, PVD~\cite{PVD}, LION~\cite{LION}, and TIGER~\cite{Tiger}—as well as leading autoregressive models, including PointGrow~\cite{PointGrow}, CanonicalVAE~\cite{pointvqvae}, and PointGPT~\cite{PointGPT}. We report results for two variants of our model, \method-s and \method-m, corresponding to small and medium parameter configurations, respectively. \textit{Note that we only compare against models with publicly available implementations.}
\textbf{Quality.} As shown in Table~\ref{tab:main_result}, \method consistently surpasses all baseline methods on both the conventional and LION data splits under the standard $2048$-point setting, establishing new SoTA generation quality. Notably, even the lightweight \method-s achieves highly competitive results. Under the dense $8192$-point setting, Table~\ref{tab:merged_dense_manyclass} (Left) further demonstrates that \method outperforms strong baselines by an increasingly larger margin. \textbf{Efficiency.} Table~\ref{tab:traing-inference-time} reports the training time, sampling time, and model size for several strong baseline models from Table~\ref{tab:main_result}, including PVD, LION, TIGER, and PointGPT. Among these, \method-s achieves the shortest training time, fastest sampling speed, and highest parameter efficiency, while still delivering competitive performance. \method-m attains state-of-the-art generation quality with the second-best efficiency, slightly trailing its lighter counterpart. Under the dense $8192$-point setting, \method-m requires only about half as many parameters as leading diffusion-based models while maintaining superior performance. Collectively, these results highlight the strong scalability potential of \method. 

\begin{figure}[htbp] 
    \centering 
    \begin{minipage}{0.45\textwidth}
        \centering 
        \includegraphics[width=\textwidth]{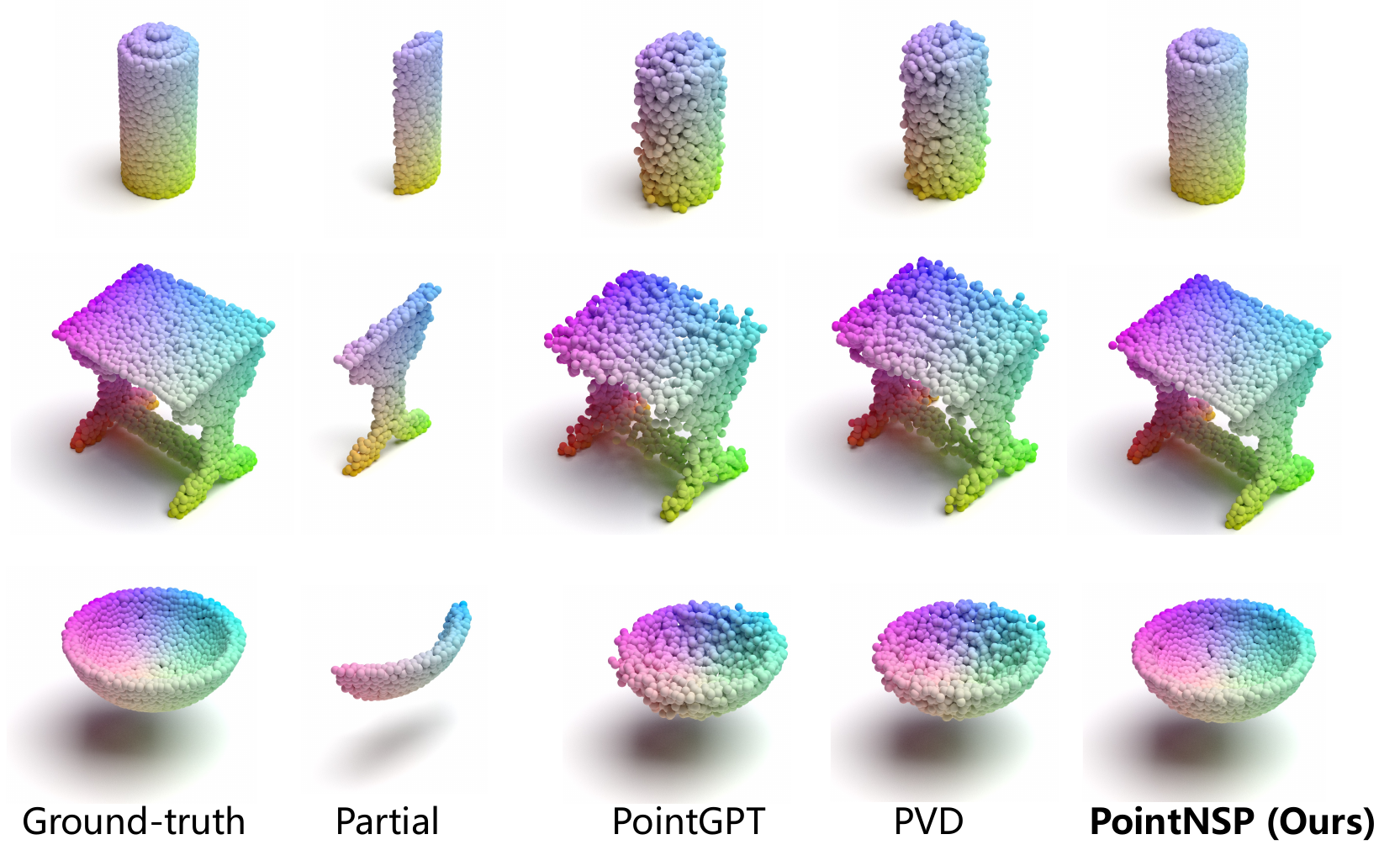} 
    \end{minipage}
    \hfill 
    \begin{minipage}{0.48\textwidth}
        \centering
        \includegraphics[width=\textwidth]{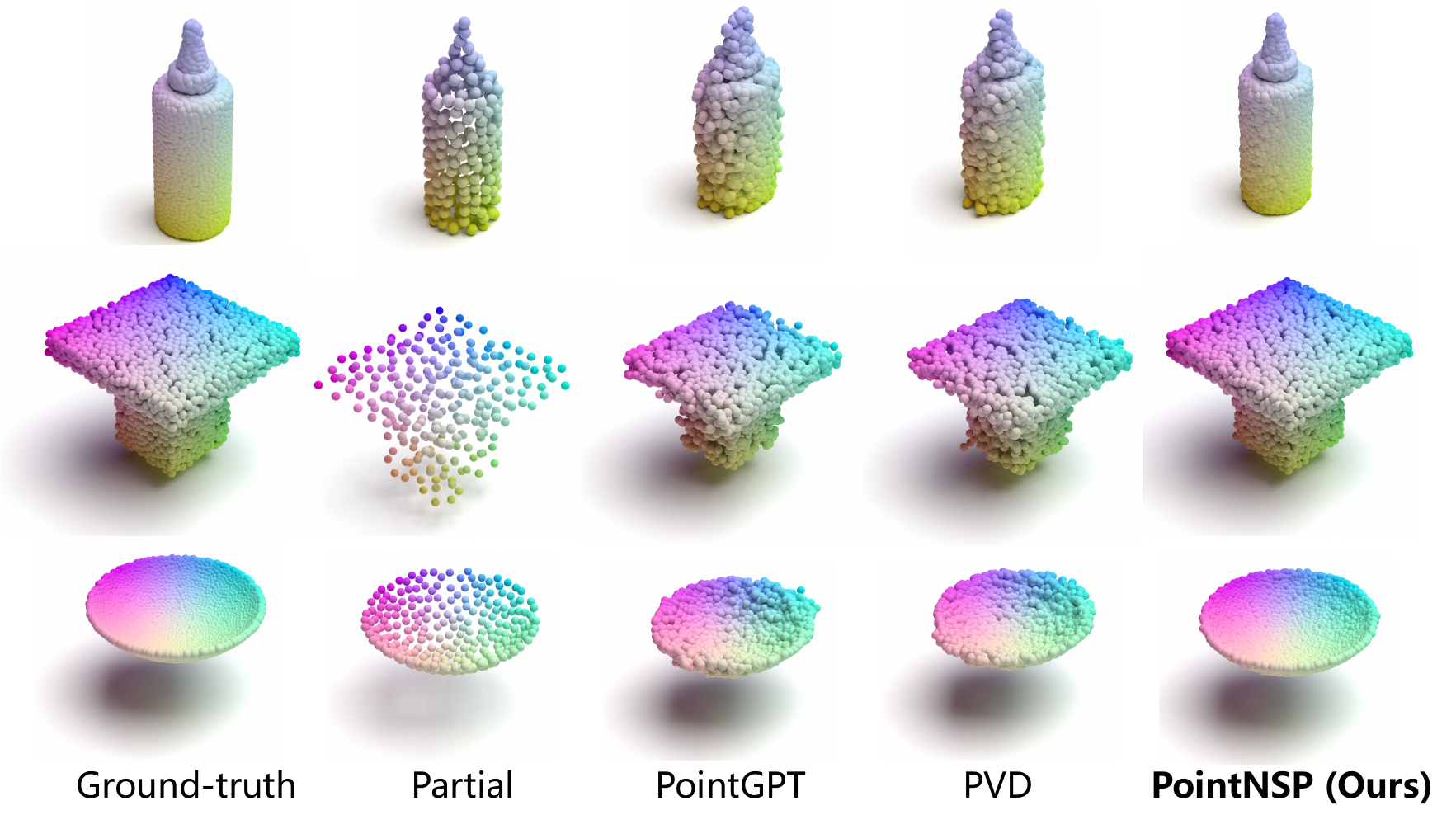}
    \end{minipage}
    
    \caption{(Left) Visualizations of point cloud completion results. (Right) Visualizations of point cloud upsampling results.}
    \label{fig:completion-upsample}
\end{figure}

\subsection{Many-Class Unconditional Generation} 

Beyond the Single-Class generation setting, we further evaluate our model on the more challenging Many-Class generation task introduced by LION~\cite{LION}, aiming to assess its ability to generalize across diverse object categories. Specifically, we train \method and selected strong baseline models without using class conditioning over $55$ distinct categories from ShapeNet. This poses a challenge for the model to capture multi-modal and structurally complex geometric patterns across categories. As shown in Table~\ref{tab:merged_dense_manyclass} (Right), \method significantly outperforms all strong baselines. We provide visualizations of diverse shape classes in Appendix~\ref{appendix:more-visual}. 

\subsection{Point Cloud Completion \& Upsampling}

We evaluate \method on two key downstream tasks: point cloud completion and upsampling. For completion, we follow the experimental setup of PVD~\citep{PVD}. For upsampling, we use a factor of $r=2$, increasing the input point clouds from $1024$ to $2048$ points. As shown in Table~\ref{tab:completion_combined}, \method consistently achieves the best performance on point cloud completion. For the upsampling task, it outperforms all selected baselines across both metrics, highlighting its effectiveness on diverse downstream applications and its potential as a foundation model. Visualization results for these two tasks are presented in Figure~\ref{fig:completion-upsample} and Appendix~\ref{appendix:more-visual}. 

\subsection{Ablation and Analysis}

\begin{table}[ht]
    \centering
    \scalebox{0.55}{
    \begin{tabular}{ccccc|cc}
    \toprule
    \multirow{2}{*}{\textbf{Position Mask}} & \multicolumn{2}{c}{\textbf{Upsampling}} & \multirow{2}{*}{\textbf{FPS Stochasticity}} & \multirow{2}{*}{\textbf{TE}} & \multirow{2}{*}{\textbf{Mean CD} $\downarrow$} & \multirow{2}{*}{\textbf{Mean EMD} $\downarrow$} \\
    \cmidrule(lr){2-3}
    & \textbf{Voxel} & \textbf{PU-Net} & & & & \\
    \midrule
    &  & \checkmark & & SE & 64.25 & 60.53\\
    \checkmark &  & \checkmark & & SE & 63.86 & 59.95\\
    \checkmark &  & \checkmark & & SE+A-PE & 62.19 & 58.23\\
    \checkmark & \checkmark &  & & SE+A-PE & 63.05 & 58.47\\
    \checkmark &  & \checkmark & & L-PE & 63.22 & 59.71\\
    \checkmark & &  \checkmark & & A-PE & 62.12 & 60.02\\
    \checkmark &  & \checkmark & \checkmark & A-PE & 61.28 & 57.32 \\ 
    \checkmark &  & \checkmark & \checkmark & SE+L-PE & 60.62 & 57.34 \\ 
    \rowcolor{green!20}
    \textbf{\checkmark} &  & \textbf{\checkmark} & \textbf{\checkmark} & \textbf{SE+A-PE} & \textbf{59.65} & \textbf{56.13} \\ 
    \bottomrule
\end{tabular}}
\caption{Training time (in GPU hours, averaged over three categories), sampling time (in seconds, averaged over samples), and model size (in millions of parameters). Ranked by generation quality on $2048$ and $8192$ settings.} 
\label{tab:ablation-studies}
\end{table}

\begin{table}[ht]
\centering
\scalebox{0.60}{
\begin{tabular}{clcc|lcc}
\Xhline{3\arrayrulewidth}
\multirow{2}{*}{\textbf{Category}} &
\multicolumn{3}{c|}{\textbf{Point Cloud Completion}} &
\multicolumn{3}{c}{\textbf{Point Cloud Upsampling}} \\
\cline{2-7}
& \textbf{Model} & \textbf{CD $\downarrow$} & \textbf{EMD $\downarrow$} &
  \textbf{Model} & \textbf{CD $\downarrow$} & \textbf{EMD $\downarrow$} \\
\Xhline{3\arrayrulewidth}

\multirow{5}{*}{Airplane} 
& SoftFlow & 40.42 & 11.98 & PVD & 73.56 & 71.65\\
& PointFlow & \underline{40.30} & 11.80 & TIGER & 71.65 & 59.94\\
& DPF-Net & 52.79 & 11.05 & PointGPT & 72.11 & 60.12\\
& PVD & 44.15 & \underline{10.30} & LION & \underline{70.41} & \underline{59.65}\\
\rowcolor{green!20}
& \textbf{\method-m (\textit{ours})} & \textbf{40.12} & \textbf{10.08} & \textbf{\method-m (\textit{ours})} & \textbf{68.89} & \textbf{58.86}\\
\cline{1-7}

\multirow{5}{*}{Chair} 
& SoftFlow & 27.86 & 32.95 & PVD & 53.81 & 64.61\\
& PointFlow & \underline{27.07} & 36.49 & TIGER & \underline{52.80} & \underline{52.98}\\
& DPF-Net & 27.63 & 33.20 & PointGPT & 53.75 & 53.21\\
& PVD & 32.11 & \underline{29.39} & LION & 53.98 & 54.33\\
\rowcolor{green!20}
& \textbf{\method-m (\textit{ours})} & \textbf{27.02} & \textbf{28.78} & \textbf{\method-m (\textit{ours})} & \textbf{52.04} & \textbf{51.03}\\
\cline{1-7}

\multirow{5}{*}{Car} 
& SoftFlow & 18.50 & 27.89 & PVD & 58.95 & 48.43\\
& PointFlow & 18.03 & 28.51 & PointGPT & 57.26 & 47.85\\
& DPF-Net & \underline{13.96} & 23.18 & TIGER & 57.90 & 48.01\\
& PVD & 17.74 & \underline{21.46} & LION & \underline{57.14} & \underline{47.56}\\
\rowcolor{green!20}
& \textbf{\method-m (\textit{ours})} & \textbf{13.84} & \textbf{20.68} & \textbf{\method-m (\textit{ours})} & \textbf{55.85} & \textbf{46.74}\\
\Xhline{3\arrayrulewidth}
\end{tabular}
}
\caption{Comparison on partial shape completion (left) and point cloud upsampling task (right).}
\label{tab:completion_combined}
\end{table}

We conduct comprehensive ablation studies to assess the impact of various architectural components and training strategies. First, we compare two upsampling strategies: \textit{voxel-based representations} and \textit{PU-Net}. Our experiments show that PU-Net consistently outperforms voxel-based upsampling, owing to its permutation-equivariant design. Second, we evaluate the effectiveness of the \textit{position-aware masking strategy}, which significantly boosts model performance. Third, we analyze the contribution of each embedding in the token embedding layer, with results highlighting the notable impact of the scale embedding. We also compare \textit{learnable positional encoding (L-PE)} with \textit{absolute PE (A-PE)}, finding that A-PE yields better performance. Finally, we examine the impact of \textit{$\operatorname{FPS}$ stochasticity}, which stems from the inherent randomness of the $\operatorname{FPS}$ algorithm. Compared to a variant with a fixed initialization that yields deterministic downsampling, our results show that incorporating FPS stochasticity consistently improves model generalization. These findings are summarized in Table~\ref{tab:ablation-studies}, with additional ablations—such as varying the number of scales—presented in Appendix~\ref{appendix:hyperparameters-reproducibility}.
\section{Conclusions}
\label{sec:conclusions}
In this work, we introduce \method, a novel autoregressive framework for high-quality 3D point cloud generation. Unlike previous approaches, \method employs a coarse-to-fine strategy that captures the multi-scale level of detail (LoD) of 3D shapes while preserving global structural coherence, rather than decomposing generation into local predictions. Our method consistently surpasses existing point cloud generative models in quality, while achieving substantial gains in parameter efficiency, training efficiency, and inference speed. Looking ahead, we aim to extend \method toward foundation-scale models trained on large-scale 3D datasets as a promising future direction.


\clearpage

{
    \small
    \bibliographystyle{ieeenat_fullname}
    \bibliography{main}
}

\clearpage
\setcounter{page}{1}
\maketitlesupplementary

\renewcommand{\thefigure}{S\arabic{figure}}
\renewcommand{\thetable}{S\arabic{table}}
\setcounter{figure}{0} 
\setcounter{table}{0}

\section{Permutation Invariance of Probability Distribution}
\label{appendix:PI} 

Here we show why the distribution $p(\mathbf{X}_{1},\mathbf{X}_{2},\dots,\mathbf{X}_{K}) = \prod_{k=1}^{K}p(\mathbf{X}_{k}|\mathbf{X}_{k-1},\dots,\mathbf{X}_{2},\mathbf{X}_{1})$ in Eq.~\ref{eq:nsp} preserves the permutation invariance  property $p(\pi(\mathbf{x}_1,\dots,\mathbf{x}_N)) = p(\mathbf{x}_1,\dots,\mathbf{x}_N), \forall \ \pi \in S_N$ in Eq.~\ref{eq:permutation-invariance}. 

By definition, the joint distribution factorizes as:
\begin{equation}
\begin{aligned}    p(\mathbf{X}_{1},\mathbf{X}_{2},\dots,\mathbf{X}_{K}) = p(\mathbf{X}_{1})p(\mathbf{X}_{2}|\mathbf{X}_{1})\\
    \dots p(\mathbf{X}_{K}|\mathbf{X}_{1},\dots \mathbf{X}_{K-1}).
\end{aligned}
\end{equation}
Consider an arbitrary permutation $\pi$ acting on the full set of points $\mathbf{X} = \bigcup_{k=1}^K \mathbf{X}_k$. This permutation can be decomposed into independent permutations within each scale:
\begin{equation}
    \pi = (\pi_{1}, \pi_{2}, \dots, \pi_{K}),\qquad  \pi_{k}\in S_{s_{k}}.
\end{equation}
Then we aim to prove that: 
\begin{equation}
    p(\pi_{1}(\mathbf{X}_{1}),\dots, \pi_{K}(\mathbf{X}_{K})) = p(\mathbf{X}_{1},\dots \mathbf{X}_{K}).
\end{equation} 
Recall that $\operatorname{FPS}$ is permutation-invariant, the resulting LoD sampling sequence $(\mathbf{X}_{1}, \dots, \mathbf{X}_{K})$ is \textit{permutation-invariant} (its inherent stochasticity operates at the set level and is independent of point ordering). The core components of \method—namely the feature encoder $\mathcal{E}(\cdot)$, $\operatorname{upsampling}$, $\operatorname{query}$, and the decoder $D(\cdot)$—are all \textit{permutation-equivariant}, ensuring that each output feature remains aligned with its corresponding input point. Therefore, the mapping between $\mathbf{X}_{k}$ and conditioning context shapes $(\mathbf{X}_{1}, \dots, \mathbf{X}_{k-1})$ remains unchanged with respect to any global permutation $\pi$. Then, for any permutation $\pi_k$ of points within $\mathbf{X}_k$: 
\begin{equation}
\begin{aligned}
    &p(\pi_{k}(\mathbf{X}_{k})|\pi_{1}(\mathbf{X}_{1}),\dots \pi_{k-1}(\mathbf{X}_{k-1})) \\
    &= p(\mathbf{X}_{k}|\mathbf{X}_{1},\dots \mathbf{X}_{k-1}).
\end{aligned}
\end{equation}
This holds for each $k = 1, \dots, K$. 
Then the joint distribution under these permutations is
\begin{equation}
\begin{aligned}
    &p(\pi_{1}(\mathbf{X}_{1}),\dots,\pi_{K}(\mathbf{X}_{K}))\\
    &= \prod_{k=1}^{K}p(\pi_{k}(\mathbf{X}_{k})|\pi_{1}(\mathbf{X}_{1}),\dots \pi_{k-1}(\mathbf{X}_{k-1})) \\ 
    &= \prod_{k=1}^{K}p(\mathbf{X}_{k}|\mathbf{X}_{k-1},\dots,\mathbf{X}_{2},\mathbf{X}_{1}) \\
    &= p(\mathbf{X}_{1},\dots,\mathbf{X}_{K}) 
\end{aligned}
\end{equation} 
\begin{figure}[ht]
    \centering
    \includegraphics[width=0.48\textwidth]{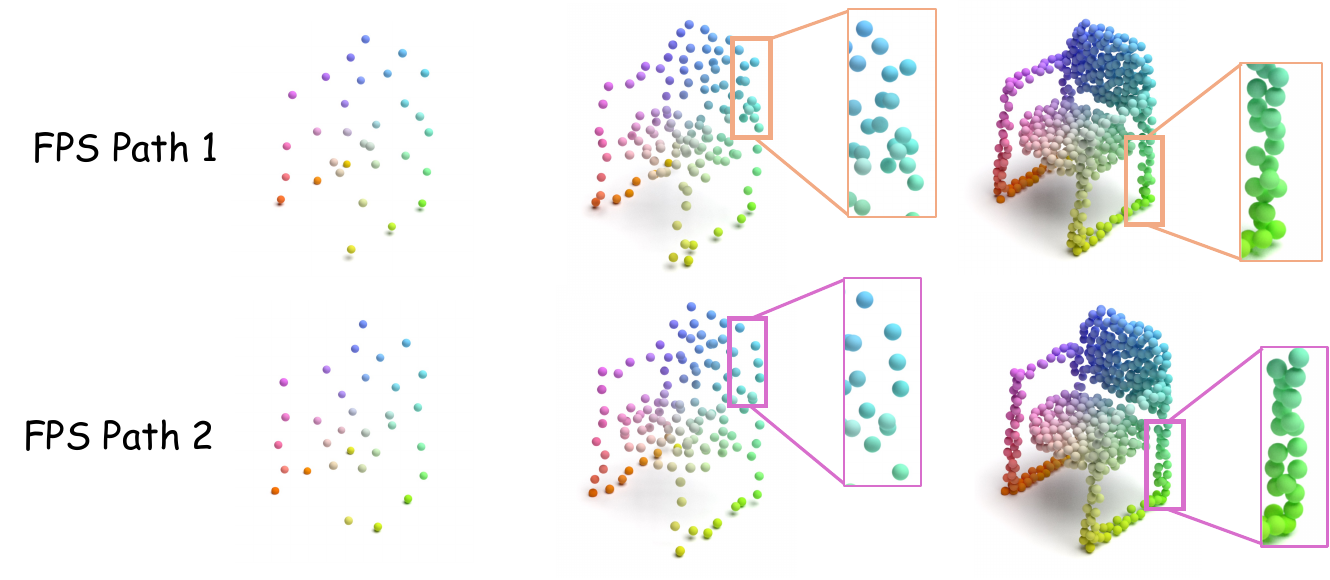}
    \caption{Illustration of different LoD sequences for a single shape. Owing to the inherent stochasticity of the sampling strategy in Eq.~\ref{eq:LoD-FPS}, our method naturally produces \textit{diverse causal LoD sequences} for the same 3D shape across training epochs. At each scale, these variations encourage \textit{broader spatial coverage} of the underlying geometry, thereby enhancing learning robustness.}
    \label{fig:FPS-path} 
\end{figure}
Consequently, the autoregressive factorization preserves permutation invariance: permuting the points within any scale does not change the resulting joint probability: 
\begin{equation}
    p(\pi(\mathbf{x}_1,\dots,\mathbf{x}_N)) = p(\mathbf{x}_1,\dots,\mathbf{x}_N), \forall \ \pi \in S_N.
\end{equation}

\section{Algorithm Details}
\label{appendix:details}

\paragraph{LoD Sequence Sampling.} This step is a core component of \method training, and we provide additional details here to eliminate any potential ambiguity. If one were to directly apply the original image-based VAR framework~\cite{VAR} to 3D point clouds, latent features for sampled LoD sequences would be obtained via
\begin{equation}
\label{eq:VAR-LoD}
    \mathbf{f}_{k} = \operatorname{FPS}(\mathbf{f}^{k-2}- \tilde{\mathbf{f}}_{k-1}), \mathbf{f}_{1} = \operatorname{FPS}(\mathbf{f}^{0}).
\end{equation}
However, this strategy is unsuitable for 3D point clouds. Applying $\operatorname{FPS}$ directly in latent space implicitly assumes that latent distances form a meaningful metric, yet these distances need not correlate with underlying geometric structure. As a result, sampling in latent space cannot reliably preserve geometric uniformity or spatial coverage. To address this, rather than iteratively applying $\operatorname{FPS}$ in latent space, we apply $\operatorname{FPS}$ in Euclidean 3D space in a \textit{fine-to-coarse} manner:
\begin{equation}
\label{eq:LoD-FPS}
    \mathbf{X}_{k-1} = \operatorname{FPS}(\mathbf{X}_{k}), \mathbf{X}_{K} = \mathbf{X}. 
\end{equation}
This produces a geometrically consistent LoD sequence $(\mathbf{X}_{1}, \dots, \mathbf{X}_{K})$ from coarse to fine, with guaranteed point correspondences $\mathbf{X}_{1}\subsetneq\dots \subsetneq\mathbf{X}_{K-1}\subsetneq \mathbf{X}_{K}$. We then obtain latent features by querying the indices mapping each $\mathbf{X}_{k}$ back to the original point set:  
\begin{equation}
    \mathbf{f}_{k} = \operatorname{query}(\mathbf{f}^{k-2} - \tilde{\mathbf{f}}_{k-1}, \mathbf{X}_{k}),\ \mathbf{f}_{1} = \operatorname{query}(\mathbf{f}^{0}, \mathbf{X}_{1}),
\end{equation}
where $\mathbf{f}^{0}=\mathcal{E}(\mathbf{X})$. This is Eq.~\ref{eq:residual-query}. Here we need to emphasize that the LoD sequence $(\mathbf{X}_{1}, \dots, \mathbf{X}_{K})$ is constructed in advance following Eq.~\ref{eq:LoD-FPS} and the Line 5 ($\mathbf{X}_{k}=\operatorname{FPS}(\mathbf{X}, s_{k})$) in Algorithm~\ref{alg:encoder} denotes that $\mathbf{X}_{k}$ with scale $s_{k}$ is obtained starting from $\mathbf{X}$. Note that it does not mean the $\mathbf{X}_{k}$ is obtained by directly applying $\operatorname{FPS}$ on the original $\mathbf{X}$ as this would break the point correspondence across scales (i.e. $\mathbf{X}_{i}\cap\mathbf{X}_{j}\not=\mathbf{X}_{i}, i<j$). Since $\operatorname{FPS}$ is inherently stochastic due to its random initialization, we can naturally obtain diverse LoD sequences for each shape $\mathbf{X}$ across training epochs. This variability is desirable, as the refinement path need not be unique; sampling multiple paths improves spatial coverage, as discussed in Section~\ref{subsec:multi-scale-LoD}. For example, for epoch $0$ and epoch $1$, we may obtain two LoD sequences for shape $\mathbf{X}$:
\begin{equation}
\label{eq:diverse-path}
    (\mathbf{X}^{0}_{1},\dots,\mathbf{X}^{0}_{K}),\ (\mathbf{X}^{1}_{1},\dots,\mathbf{X}^{1}_{K}).
\end{equation}
At each scale, $\mathbf{X}^{0}_{k}$ and $\mathbf{X}^{1}_{k}$ represent the same underlying surface but with different point coverages, which helps improve learning robustness and generalization. We illustrate this LoD sequence sampling issue in Figure~\ref{fig:FPS-path}. 

\paragraph{Coordinate-based Positional Encoding.} We adopt the absolute positional encoding strategy purely based on 3D coordinates used in TIGER~\citep{Tiger}. Based on our experiments, we find the Base $\lambda$ Position Encoding (B$\lambda$PE) performs better and here we present its formula:
\begin{align}
    p &= \lambda^{2} * z_{i} + \lambda * y_{i} + x_{i} \\
    & \mathbf{P}_{k}(p, 2i) = \sin(\frac{p}{10000^{\frac{2i}{D}}})\\
    & \mathbf{P}_{k}(p, 2i+1) = \cos(\frac{p}{10000^{\frac{2i}{D}}}), 
\end{align} 
where $\mathbf{x}_{i} = \mathbf{X}_{k}[i] = (x_{i}, y_{i}, z_{i})\in \mathbb{R}^{3}$, $p$ is a polynomial expression with hyperparameter coefficient $\lambda$. We set $\lambda=1000$ following the setting in TIGER~\citep{Tiger}, which means this preserves three decimal places of precision. Here $\mathbf{P}_{k}\in \mathbb{R}^{s_{k}\times d}$ denotes the positional embedding of all tokens within the scale $k$. In short, we apply the B$\lambda$PE embedding strategy scale-by-scale. 

\paragraph{Intra-Scale Token Embedding.} Instead of simply adding all token embeddings together, the real implementation is analogous to Llama~\citep{Llama3} by adding positional embedding and scale embedding to query and key vectors. Specifically, we retrieve the codebook embedding $\mathbf{z}_{k}$ for each scale token $q_{k}$ and then upsampled from input scale to output scale $(s_{k}\rightarrow s_{k+1})$: $\mathbf{z}_{k} = \operatorname{upsampling}(\mathbf{z}_{k}) \in \mathbb{R}^{s_{k+1}\times d}$ following the Eq.~\ref{eq:up-sample}. Positional embedding $\mathbf{p}^{i}_{k} = \mathbf{P}_{k}[i]$ for each token $q_{k}^{i}$ is derived with from the decoded intermediate structure $\mathbf{X}_{k}$ as described in Eq.~\ref{eq:intermediate}. During inference stage, the intermediate structure $\hat{\mathbf{X}}_{k}$ is decoded using the predicted token $\hat{q}_{k}$ instead. Additionally, the model needs to know which scale that each token belongs to. Therefore, $\mathbf{s}_{k}$ is a simple one-hot embedding $\mathbf{s}_{k} = \operatorname{one-hot-embedding}(k)$ out of total $K$ scales. Tokens from the same scale $k$ share the same scale embedding $\mathbf{s}_{k}$ ($\mathbf{s}^{i}_{k}=\mathbf{s}^{j}_{k}$ for $q_{k}^{i}$, $q_{k}^{j}$). For all input scale tokens, we add both the positional embedding $\mathbf{p}^{i}_{k}$ and the scale embedding $\mathbf{s}^{i}_{k}$ to the query $\mathbf{u}_{k}^{i}$ and key vectors $\mathbf{v}_{k}^{i}$ derived in the attention mechanism: 
\begin{equation}
\label{eq:token_embed}
    \mathbf{u}_{k}^{i} = \mathbf{W}_{\mathbf{U}}\mathbf{z}^{i}_{k} + \mathbf{p}^{i}_{k} + \mathbf{s}^{i}_{k}, \mathbf{v}_{k}^{i} = \mathbf{W}_{\mathbf{V}}\mathbf{z}^{i}_{k} + \mathbf{p}^{i}_{k} + \mathbf{s}^{i}_{k}, 
\end{equation} 
where $\mathbf{W}_{\mathbf{U}}$ and $\mathbf{W}_{\mathbf{V}}$ are projection matrices for queries and keys respectively. Together with value vectors, these vectors will be fed to the block-wise causal transformer for next-scale token prediction. 

\section{Hyperparameters \& Reproducibility Settings}
\label{appendix:hyperparameters-reproducibility}

\paragraph{Hyperparameters \& Reproducibility Settings.} Specifically, we set the learning rate $3e^{-4}$ and the batch-size 32. We perform all the experiments on a workstation with Intel Xeon Gold 6154 CPU (3.00GHz) and 8 NVIDIA Tesla V100 (32GB) GPUs. We use an AdamW optimizer with an initial learning rate of $10^{-4}$ for VQVAE training and $10^{-3}$ for autoregressive transformer training respectively. For upsampling and completion experiments, we follow the experimental settings of PVD~\cite{PVD}. For many-class generation, we mainly inherit the experimental setting from LION~\cite{LION}. 

\begin{table}[ht]
\centering
\begin{tabular}{c|c}
\hline
Hyperparameter & \textbf{Value} \\ \hline
\# PVCNN layers & 4 \\ \hline
\# PVCNN hidden dimension & 1024  \\ \hline
\# PVCNN voxel grid size & 32  \\ \hline
\# MLP layers & 6 \\ \hline
\# Attention dimension & 1024 \\ \hline
\# Attention head & 32 \\ \hline
Optimizer & AdamW \\ \hline
Weight Decay & 0.01 \\ \hline
LR Schedule & Cosine \\ \hline
\end{tabular}
\end{table}

\begin{figure}[h]
    \centering
    \includegraphics[width=0.9\linewidth]{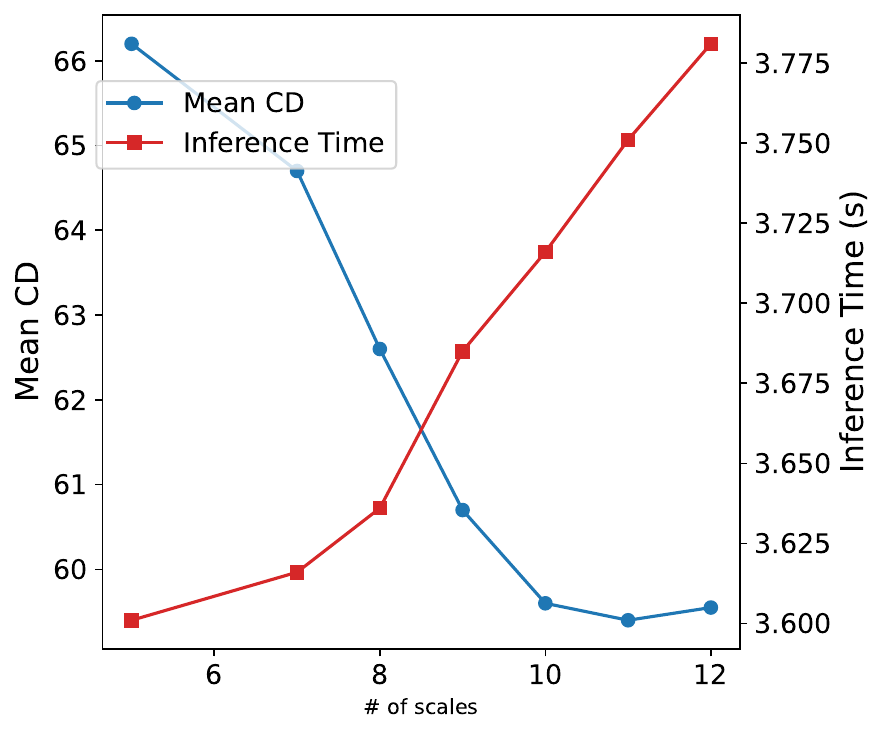}
    \caption{The effect of number of scales on the overall performance of \method.}
    \label{fig:scales}
\end{figure}

\paragraph{The effect of \# scales $K$.} Figure~\ref{fig:scales} illustrates the impact of the total number of scales on the overall performance of \method. As the number of scales increases, \method's performance improves accordingly. However, beyond $K = 11$ scales, no further performance gains are observed. We hypothesize that additional scales may require higher point cloud densities to be effective. Moreover, increasing the number of scales naturally leads to longer sampling times.

\section{More Experimental Results}
\label{appendix:more-exp-results}

Due to page limitations, the main paper reports only the strongest baseline methods in the primary comparison table. Here, we provide a more comprehensive evaluation that additionally includes 1-GAN~\cite{1-GAN} (GAN-based), PointFlow~\cite{PointFlow}, DPF-Net~\cite{DPF-Net}, SoftFlow~\cite{SoftFlow} (normalizing flow–based), SetVAE~\cite{setvae} (VAE-based), PVD-DDIM~\cite{DDIM} (diffusion models with advanced samplers), PSF~\cite{PSF} (flow-matching–based), and DIT-3D~\cite{DIT-3D}. We include only methods with publicly available implementations or those reporting explicit quantitative results in their original papers. Please see Table~\ref{tab:complete_main_result} for comprehensive comparisons.

Since ShapeNet categories are highly imbalanced, prior works typically report results only on the three largest categories—airplane, chair, and car—while the remaining categories contain significantly fewer samples. To provide a more comprehensive evaluation, we additionally report performance on three smaller categories: table, sofa, and lamp. As shown in Table~\ref{tab:per_category}, \method consistently and substantially outperforms all baseline methods, demonstrating its strong learning efficiency \textit{under limited data conditions}.

\begin{table}[htbp]
\centering
\small
\setlength{\tabcolsep}{10pt}
\resizebox{0.48\textwidth}{!}{
\begin{tabular}{lccc}
\toprule
\multirow{2}{*}{\textbf{Model}}  & \textbf{Table}   & \textbf{Sofa}    & \textbf{Lamp} \\
\cmidrule(lr){2-2} \cmidrule(lr){3-3} \cmidrule(lr){4-4}
    & CD / EMD     & CD / EMD       & CD / EMD \\
\midrule
PointFlow~\cite{PointFlow}   & 92.3 / 92.8 & 85.6 / 88.4 & 88.7 / 98.5 \\
PointGPT~\cite{PointGPT}   & 91.5 / 90.9 & 83.8 / \underline{88.4} & 88.7 / 98.5 \\
ShapeGF~\cite{ShapeGF}       & 83.7 / 81.5 & 79.2 / 81.3 & 86.4 / 86.8 \\
PVD~\cite{PVD}              & 76.8 / 80.2  & 73.5 / 96.7  & 89.1 / 89.4 \\
LION~\cite{LION}          & \underline{68.4} / \underline{78.7} & \underline{67.9} / 90.2 & \underline{81.3} / \underline{82.6} \\
\textbf{\method-m}             & \textbf{61.2} / \textbf{76.5} & \textbf{60.8} / \textbf{83.7} & \textbf{73.9} / \textbf{74.8} \\
\bottomrule
\end{tabular}
}
\caption{Per-category generation quality on ShapeNet. We report 1-NNA CD/EMD ($\downarrow$) on \textbf{Table}, \textbf{Sofa}, and \textbf{Lamp}. Best in \textbf{bold}, second best \underline{underlined}.}
\label{tab:per_category}
\end{table}

In addition to the 55-class generation setting, we also evaluate the 13-class setting used in LION~\cite{LION}. The results, presented in Table~\ref{tab:13-class}, show that \method achieves state-of-the-art generation quality.

\begin{table}[htbp]
\centering
    \begin{tabular}{lcc}
    \toprule
    \textbf{Model}              & \textbf{CD} $\downarrow$ & \textbf{EMD} $\downarrow$ \\
    \midrule
    TreeGAN~\cite{TreeGAN}     & 96.80          & 96.60          \\
    PointFlow~\cite{PointFlow}   & 63.25          & 66.05          \\
    ShapeGF~\cite{ShapeGF}      & 55.65          & 59.00          \\
    SetVAE~\cite{setvae}         & 79.25          & 95.25          \\
    PDGN~\cite{PDGN}         & 71.05          & 86.00          \\
    DPF-Net~\cite{DPF-Net}      & 67.10          & 64.75          \\
    DPM~\cite{DPM}     & 62.30          & 86.50          \\
    PVD~\cite{PVD}                & 58.65          & 57.85          \\
    LION~\cite{LION}                & \underline{55.52}          & \underline{53.89}         \\
    \midrule
    \textbf{\method-m}                & \textbf{54.70} & \textbf{52.82} \\
    \bottomrule
    \end{tabular}
\caption{Generation results (1-NNA $\downarrow$) trained jointly on 13 classes of ShapeNet-vol.}
\label{tab:13-class}
\end{table}

\begin{table*}[ht]
    \vspace{-3em}
    \caption{The \emph{Performance} (1-NNA) is evaluated based on single-class generation. The second block specifies the types of generative models used in each study. The best performance is highlighted in \textbf{{bold}}, while the second-best performance is underlined. Performance is reported on two dataset splits: the top corresponds to the random split, and the bottom corresponds to the LION split.}
    \centering
     \resizebox{0.99\textwidth}{!}{
     \begin{tabular}{l|c|cc|cc|cc|c|c}
\Xhline{3\arrayrulewidth} 
      \multirow{2}{*}{\textbf{Model}} & \multirow{2}{*}{\textbf{Generative Model}} & \multicolumn{2}{c|}{\textbf{Airplane}} & \multicolumn{2}{c|}{\textbf{Chair}} & \multicolumn{2}{c|}{\textbf{Car}} & \multirow{2}{*}{\textbf{Mean CD $\downarrow$}} & \multirow{2}{*}{\textbf{Mean EMD $\downarrow$}}\\
      \cline{3-8}
      & & CD $\downarrow$ & EMD $\downarrow$ & CD $\downarrow$ & EMD $\downarrow$ & CD $\downarrow$ & EMD $\downarrow$ & & \\
    \Xhline{1.5\arrayrulewidth} 
      1-GAN~\cite{1-GAN} & GAN & 87.30 & 93.95 & 68.58 & 83.84 & 66.49 & 88.78 & 74.12 & 88.86\\
       PointFlow~\citep{PointFlow} & Normalizing Flow & 75.68 & 70.74 & 62.84 & 60.57& 58.10 & 56.25 & 65.54 & 62.52\\
       DPF-Net~\citep{DPF-Net} & Normalizing Flow & 75.18 & 65.55 & 62.00 & 58.53& 62.35  &54.48 & 66.51 & 59.52\\
       SoftFlow~\citep{SoftFlow} & Normalizing Flow& 76.05 & 65.80 & 59.21 & 60.05& 64.77 & 60.09 & 66.67 & 61.98\\
       SetVAE~\citep{setvae} & VAE &75.31 & 77.65 & 58.76 & 61.48 & 59.66 & 61.48 & 64.58 & 66.87\\
       ShapeGF~\citep{ShapeGF} & Diffusion & 80.00 & 76.17 & 68.96  & 65.48 & 63.20 & 56.53 & 70.72 & 66.06\\
       DPM~\citep{DPM} & Diffusion & 76.42 & 86.91 & 60.05 & 74.77 & 68.89 & 79.97 & 68.45 & 80.55\\
       PVD-DDIM~\citep{DDIM} & Diffusion & 76.21 & 69.84 & 61.54 & 57.73 &  60.95 & 59.35 & 66.23 & 62.31\\
       PSF~\citep{PSF} & Diffusion & 74.45 & 67.54 & 58.92 & 54.45 & 57.19 & 56.07 & 62.41 & 57.20 \\
       PVD~\citep{PVD} & Diffusion & 73.82 & 64.81 & 56.26 & 53.32 & 54.55 & 53.83 & 61.54 & 57.32 \\
       LION~\citep{LION} & Diffusion & 72.99 & 64.21 & 55.67 & 53.82 & 53.47 & 53.21 & 61.75 & 57.59\\
       DIT-3D~\citep{DIT-3D} & Diffusion & - & - & \underline{54.58} & 53.21 & - & - & - & -\\
       TIGER~\citep{Tiger} & Diffusion & 73.02 & 64.10 & 55.15 & 53.18 & 53.21 & 53.95 & 60.46 & 57.08\\ 
        PointGrow~\citep{PointGrow} & Autoregressive  & 82.20 & 78.54 & 63.14 & 61.87 & 67.56 & 65.89 & 70.96 & 68.77\\
        CanonicalVAE~\citep{pointvqvae} & Autoregressive & 80.15 & 76.27 & 62.78 & 61.05 & 63.23 & 61.56 & 68.72 & 66.29\\
        PointGPT~\citep{PointGPT} & Autoregressive & 74.85 & 65.61 & 57.24 & 55.01 & 55.91 & 54.24 & 63.44 & 62.24\\ 
        \method-s (\textit{ours}) & Autoregressive & \underline{72.92} & \underline{63.98} & 54.89 & \underline{53.02} & \underline{52.86} & \underline{52.07} & \underline{60.22} & \underline{56.36}\\
        \rowcolor{green!20}
        \textbf{\method-m (\textit{ours})} & \textbf{Autoregressive} & \textbf{72.24} & \textbf{63.69} & \textbf{54.54} & \textbf{52.85} & \textbf{52.17} & \textbf{51.85} & \textbf{59.65} & \textbf{56.13}\\
 
 \midrule 
    LION & Diffusion & 67.41 & 61.23 & \underline{53.70} & 52.34 & \underline{53.41} & 51.14 & \underline{58.17} & 54.90\\
    TIGER & Diffusion & 67.21 & 56.26 & 54.32 & 51.71 & 54.12 & 50.24 & 58.55 & 52.74\\
    \method-s (\textit{ours}) & Autoregressive & \underline{67.15} & \underline{56.12} & 54.22 & \underline{51.19} & 53.98 & \underline{50.15} & 58.45 & \underline{52.49} \\
    \rowcolor{green!20}
    \textbf{\method-m (\textit{ours})} & \textbf{Autoregressive} & \textbf{66.98} & \textbf{56.05} & \textbf{54.01} & \textbf{53.76} & \textbf{53.12} & \textbf{50.09} & \textbf{58.04} & \textbf{52.30}\\
    \Xhline{3\arrayrulewidth} 
    \end{tabular}
    }
    \vspace{-0.5em}
    \label{tab:complete_main_result}
\end{table*} 

\section{More Visualization Results} 
\label{appendix:more-visual}

\begin{figure*}[ht]
    \centering
    \includegraphics[width=1\textwidth]{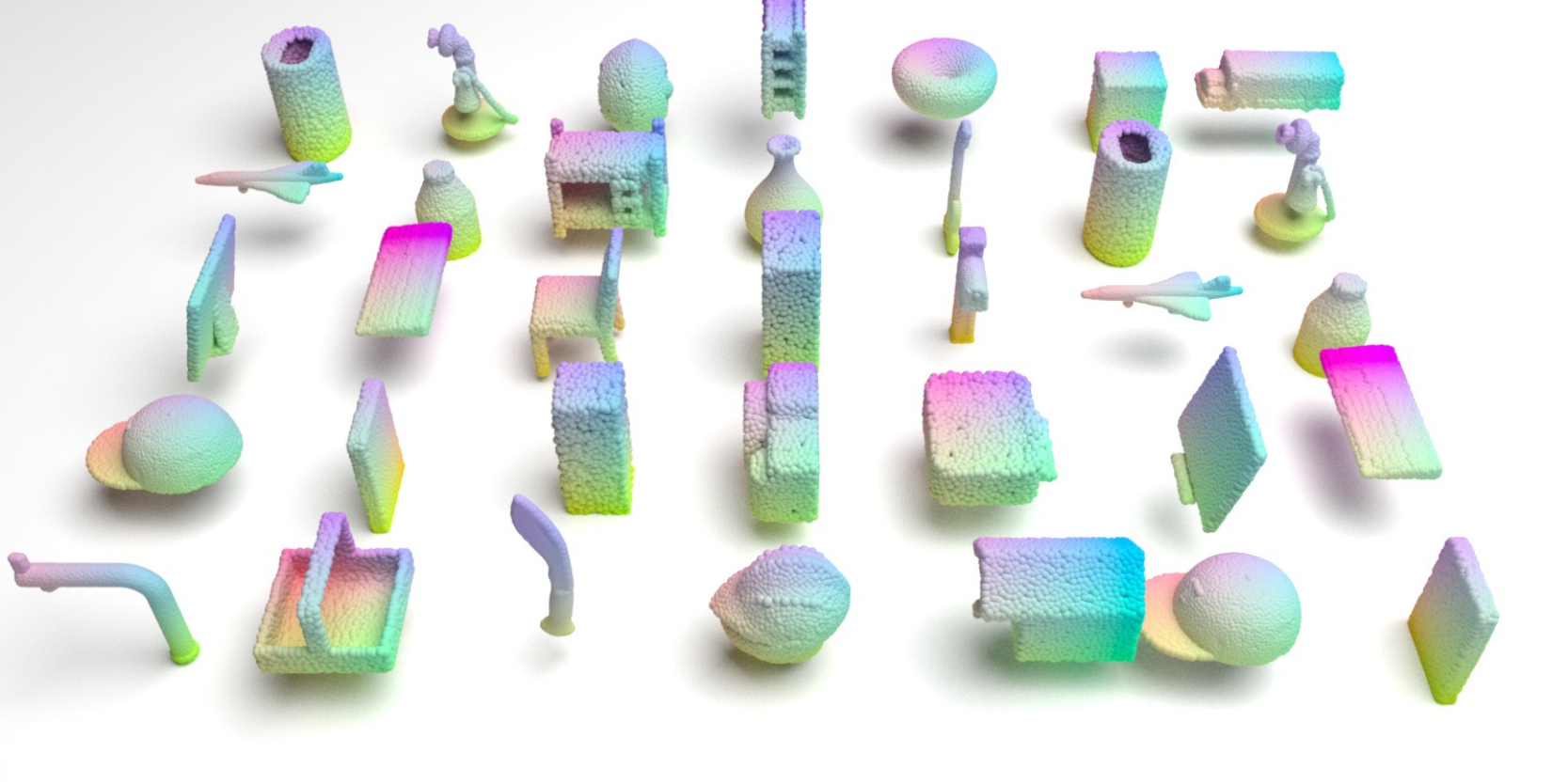}
    \caption{Generated shapes from the \method model trained on ShapeNet’s other categories.}
    \label{fig:visualization-1}
\end{figure*}

\begin{figure*}[ht]
    \centering
    \includegraphics[width=1\textwidth]{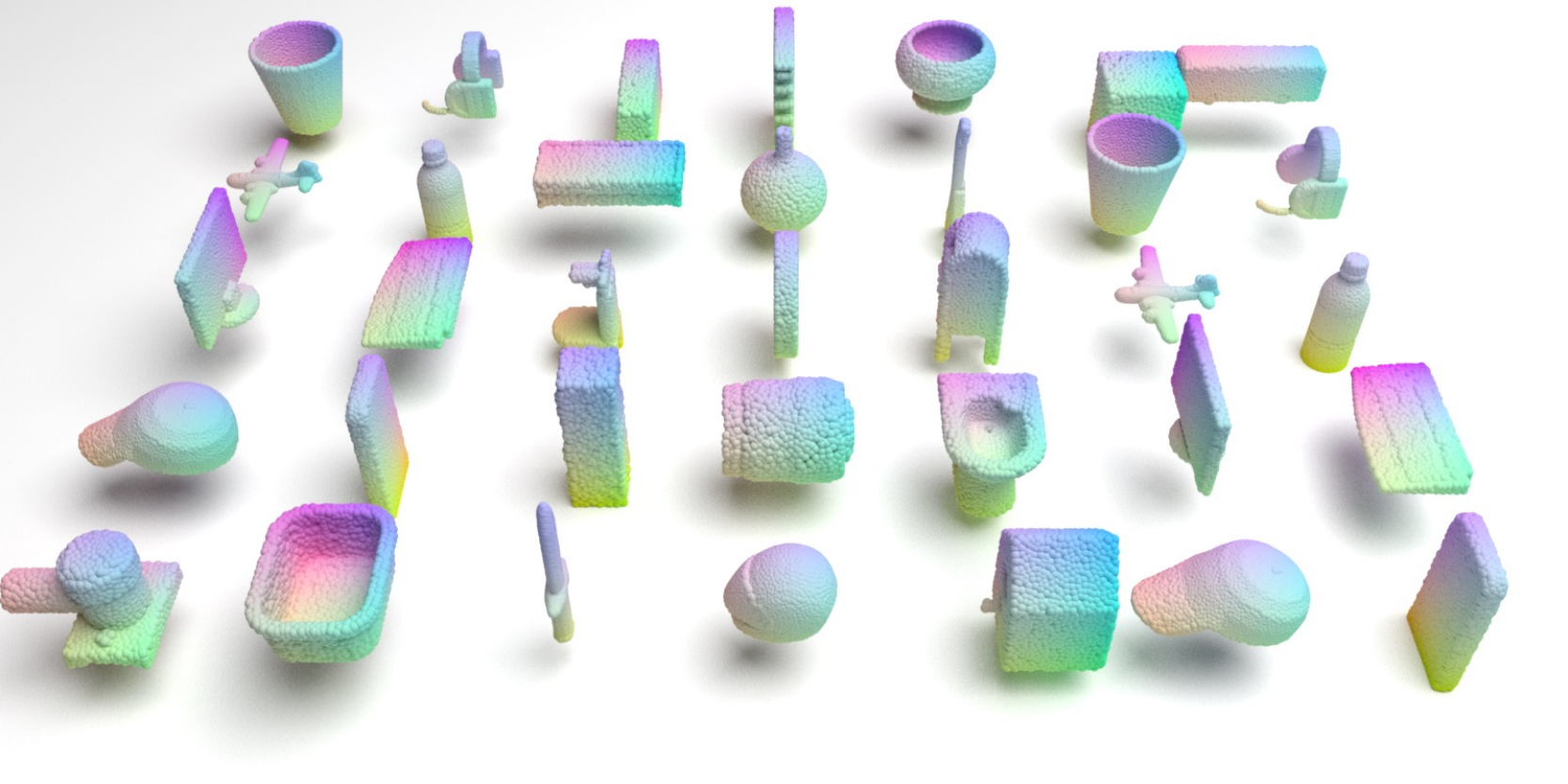}
    \caption{Generated shapes from the \method model trained on ShapeNet’s other categories.}
    \label{fig:visualization-2}
\end{figure*}

\begin{figure*}[t]
    \centering
    \includegraphics[width=1\textwidth]{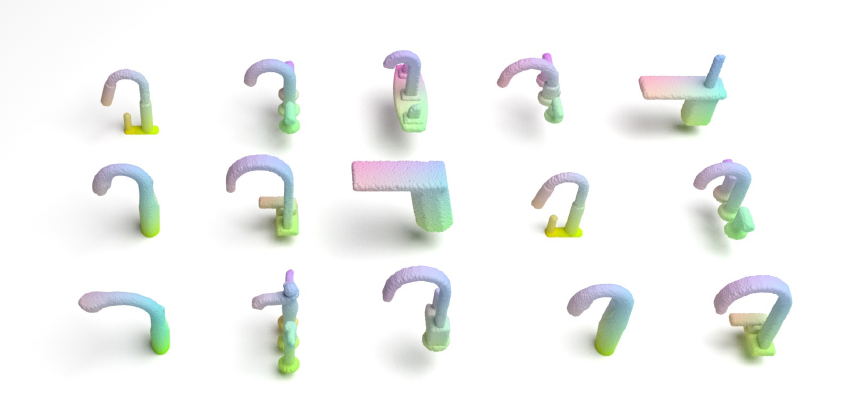}
    \caption{Generated single-class shapes from the \method model trained on ShapeNet’s other categories.}
    \label{fig:visualization-sample-1}
\end{figure*}

\begin{figure*}[t]
    \centering
    \includegraphics[width=1\textwidth]{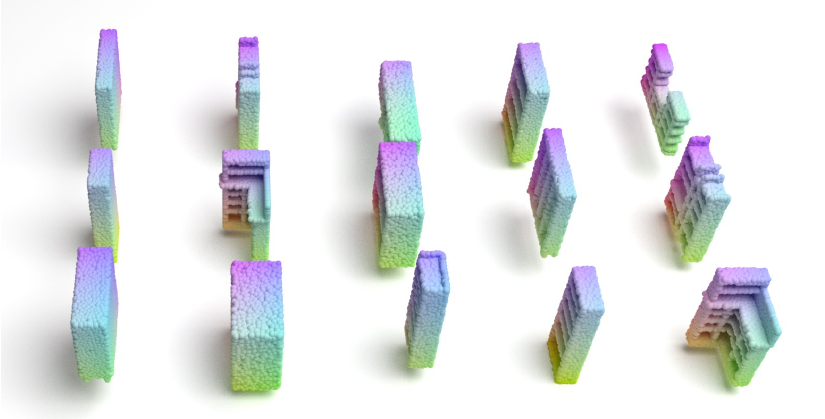}
    \caption{Generated single-class shapes from the \method model trained on ShapeNet’s other categories.}
    \label{fig:visualization-sample-2}
\end{figure*}

\begin{figure*}[t]
    \centering
    \includegraphics[width=1\textwidth]{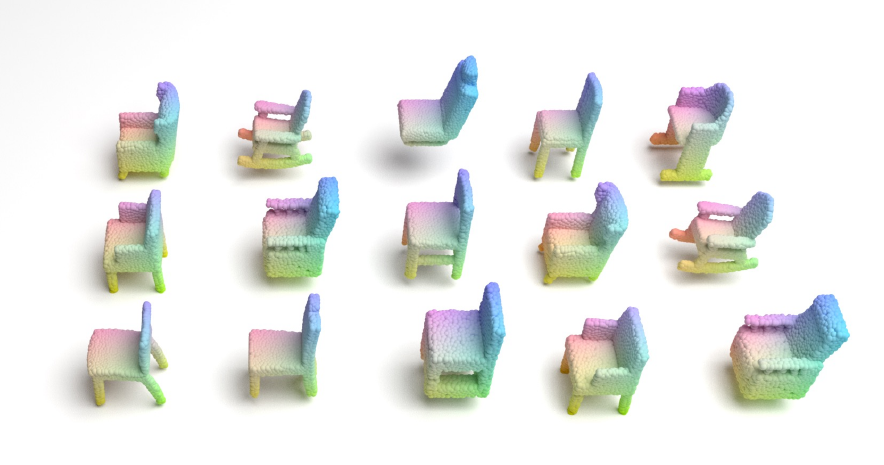}
    \caption{Generated single-class shapes from the \method model trained on ShapeNet’s other categories.}
    \label{fig:visualization-sample-3}
\end{figure*}

\begin{figure*}[t]
    \centering
    \includegraphics[width=1\textwidth]{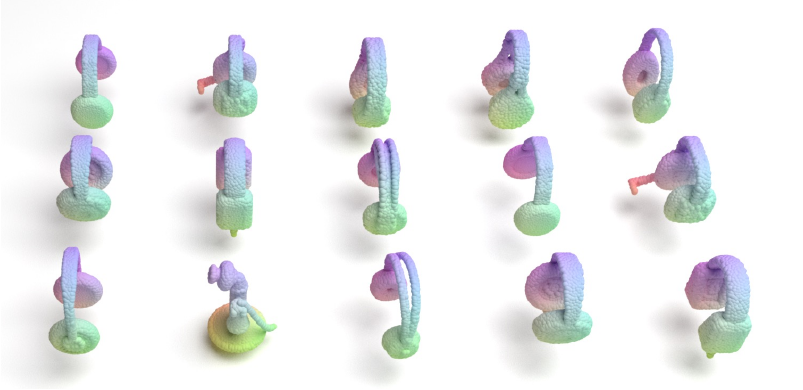}
    \caption{Generated single-class shapes from the \method model trained on ShapeNet’s other categories.}
    \label{fig:visualization-sample-4}
\end{figure*}

\begin{figure*}[t]
    \centering
    \includegraphics[width=1\textwidth]{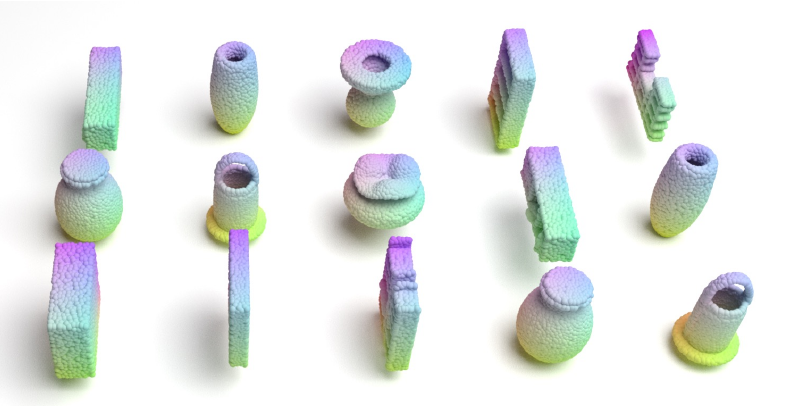}
    \caption{Generated single-class shapes from the \method model trained on ShapeNet’s other categories.}
    \label{fig:visualization-sample-5}
\end{figure*}

\begin{figure*}[t]
    \centering
    \includegraphics[width=1\textwidth]{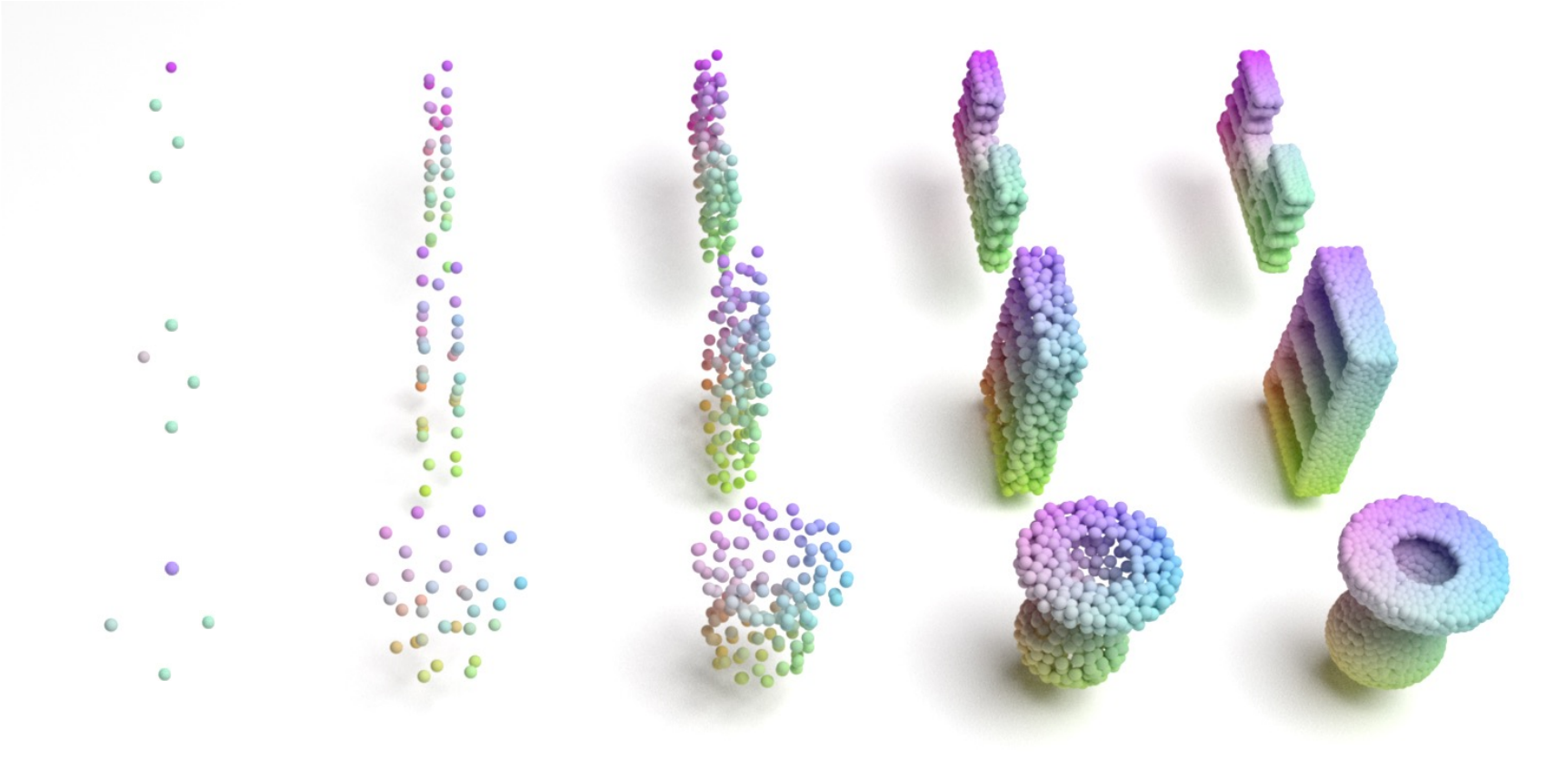}
    \caption{Illustration of multi-scale point cloud generation from the \method model.}
    \label{fig:visualization-process-1}
\end{figure*}

\begin{figure*}[t]
    \centering
    \includegraphics[width=1\textwidth]{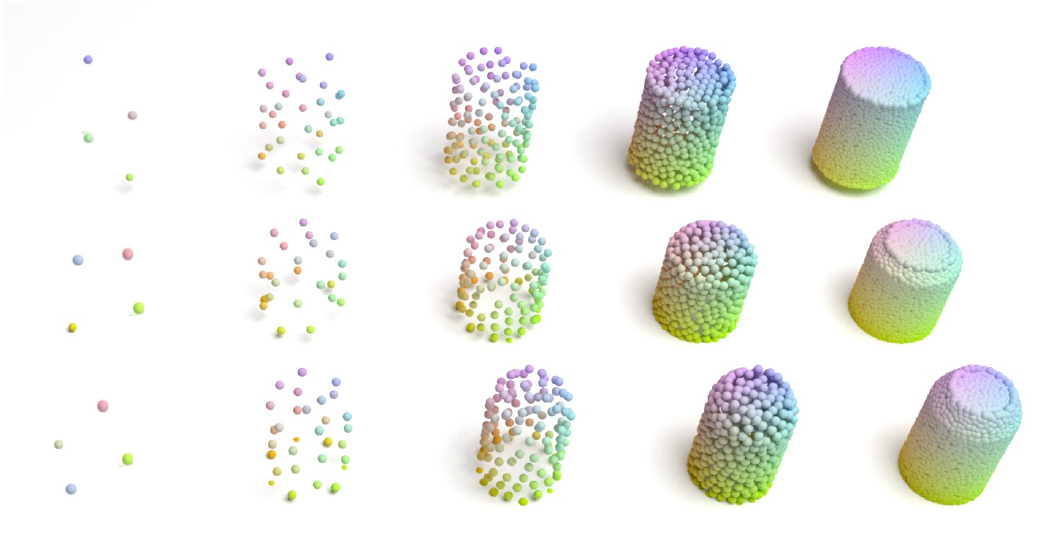}
    \caption{Illustration of multi-scale point cloud generation from the \method model.}
    \label{fig:visualization-process-2}
\end{figure*}

\begin{figure*}[t]
    \centering
    \includegraphics[width=0.95\textwidth]{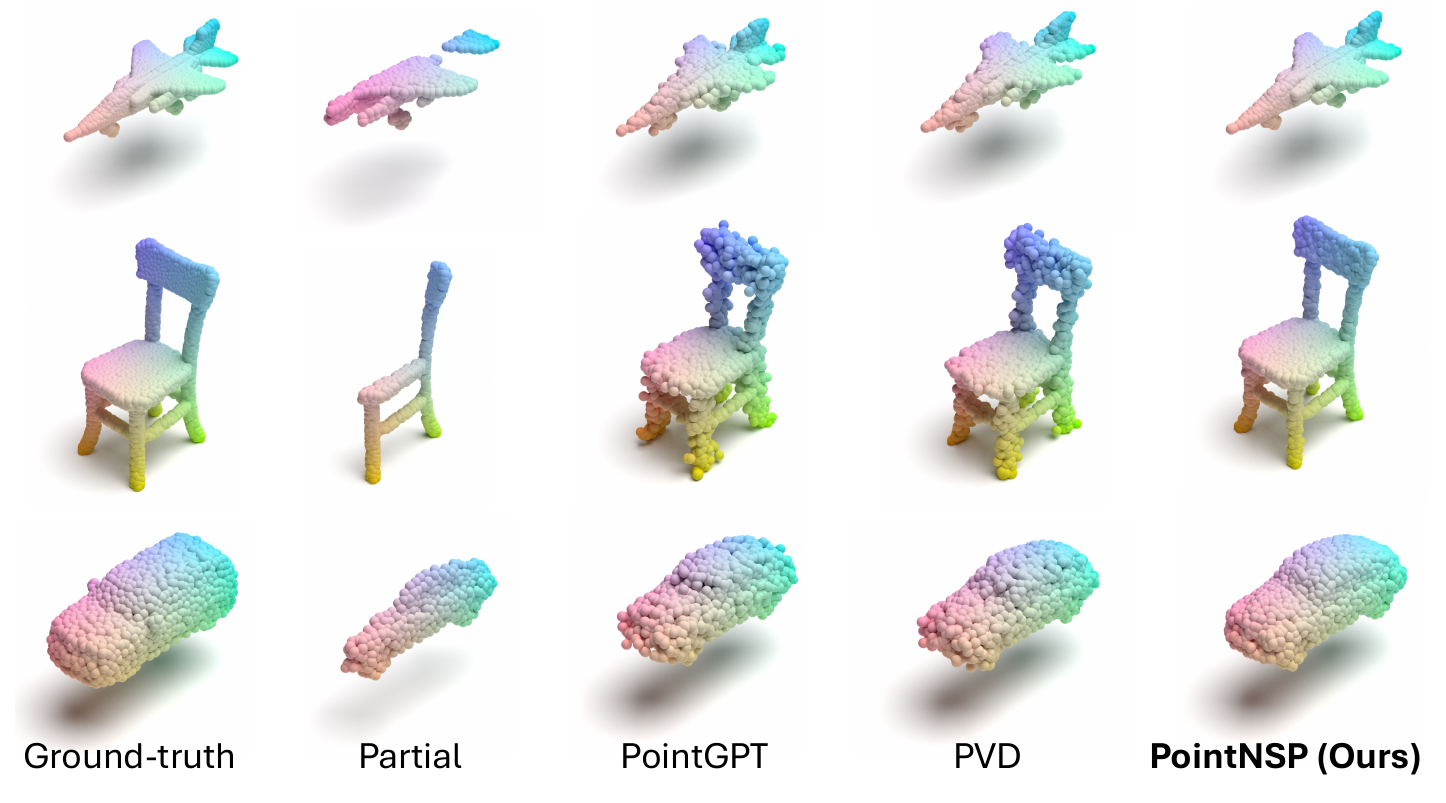}
    \caption{Visualizations of more point cloud completion results.}
    \label{figure:more-class-completion} 
\end{figure*}

\begin{figure*}[t]
    \centering
    \includegraphics[width=0.95\textwidth]{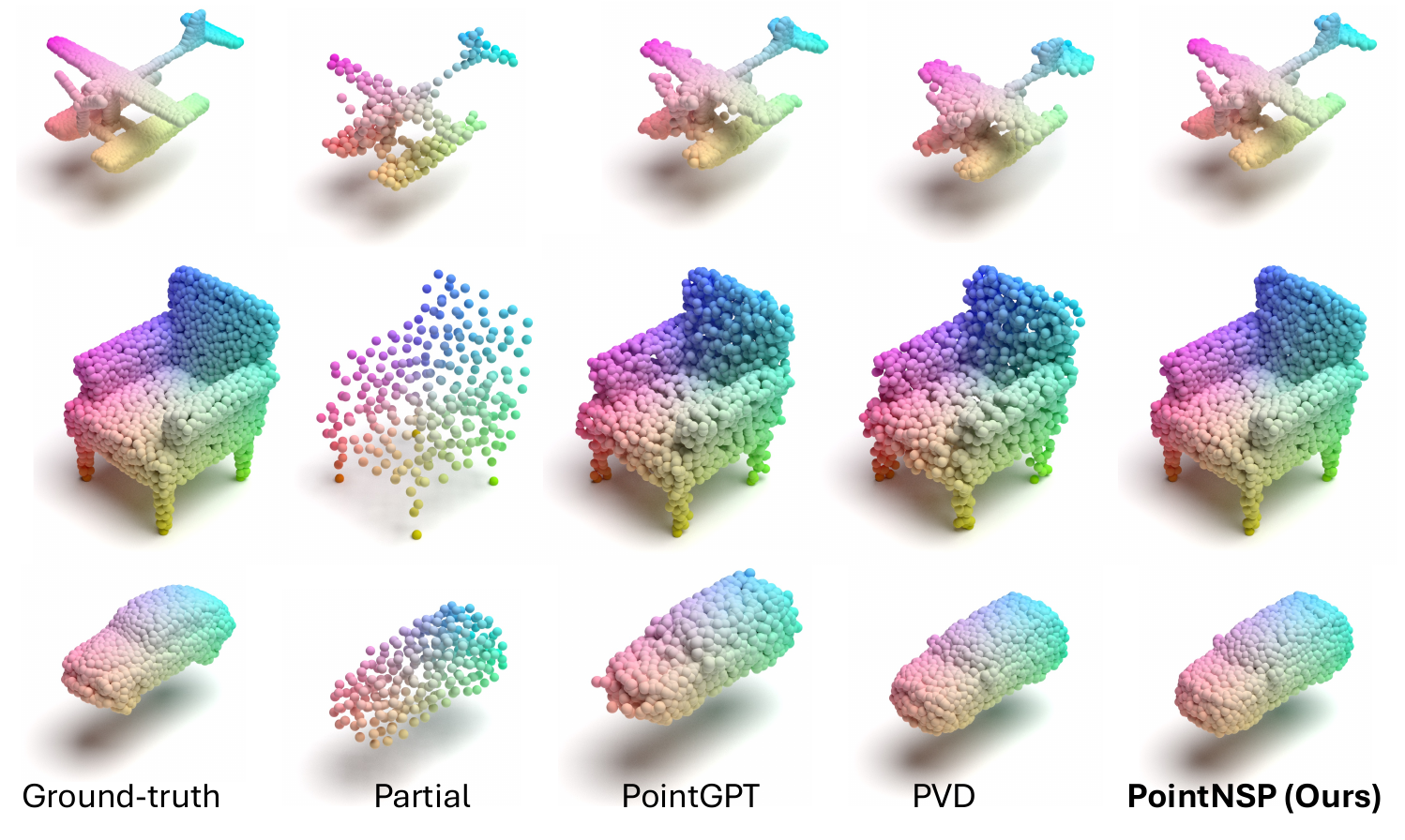}
    \caption{Visualizations of more point cloud upsampling results.}
    \label{figure:more-class-upsampling} 
\end{figure*}

We showcase diverse 3D point clouds generated by \method across a wide variety of shapes (Figures~\ref{fig:visualization-1} and \ref{fig:visualization-2}). Additional single-class generation results for five categories are provided in Figures~\ref{fig:visualization-sample-1}–\ref{fig:visualization-sample-5}. We further illustrate the multi-scale sequential generation process in Figures~\ref{fig:visualization-process-1}–\ref{fig:visualization-process-2}, and present more examples of point cloud completion and upsampling in Figures~\ref{figure:more-class-upsampling} and \ref{figure:more-class-completion}.

\section{Other Related Works}
\label{appendix:more-related-works}

\paragraph{Point Cloud Upsampling.} Point cloud upsampling is a crucial process in 3D modeling, aimed at increasing the resolution of low-resolution 3D point clouds. PU-Net~\cite{PU-net} pioneered the use of deep neural networks for this task, laying the foundation for subsequent advancements. Models such as PU-GCN~\cite{PU-GCN} and PU-Transformer~\cite{PU-transformer} have further refined point cloud feature extraction by leveraging graph convolutional networks and transformer networks, respectively. Additionally, approaches like Dis-PU~\cite{DisPU}, PU-EVA~\cite{PU-EVA}, and MPU~\cite{MPU} have enhanced the PU-Net pipeline by incorporating cascading refinement architectures. Other methods, such as PUGeo-Net~\cite{PUGeoNet}, NePs~\cite{NeuralPoints}, and MAFU~\cite{MAFU}, employ local geometry projections into 2D space to model the underlying 3D surface. More recent approaches have reframed upsampling as a generation task. For instance, PU-GAN~\cite{PU-GAN} and PUFA-GAN~\cite{PUFA-GAN} leverage generative adversarial networks (GANs) to produce high-resolution point clouds. Grad-PU~\cite{Grad-PU} first generates coarse dense point clouds through nearest-point interpolation and then refines them iteratively using diffusion models. In contrast, PUDM~\cite{PUDM} directly utilizes conditional diffusion models, treating sparse point clouds as input conditions for generating denser outputs. In this work, our generative model, \method, incorporates upsampling networks in both two training stages, making it well-suited for enhancing downstream point cloud upsampling tasks.  

\section{Limitations \& Broader Impact}

\method does not exhibit major limitations, though a primary challenge lies in learning high-quality multi-scale codebook embeddings for 3D point cloud representations, particularly in avoiding codebook collapse. Several promising directions arise for future work. One avenue is scaling generation toward ultra-dense point clouds (e.g., $10$k–$100$k points), which could subsequently be converted into high-fidelity meshes. Another is enabling fine-grained control over local geometric structures, a capability crucial for practical deployment. Although this work does not present immediate societal risks, potential misuse for generating harmful 3D content warrants careful consideration by the broader community. From a research standpoint, \method represents a significant contribution to both the generative modeling and 3D point cloud communities. 

\end{document}